\documentclass[11pt]{article}

\usepackage[a4paper, margin=1in]{geometry}
\usepackage[utf8]{inputenc}
\usepackage[T1]{fontenc}
\usepackage{lmodern}

\usepackage[blocks]{authblk}

\usepackage{natbib}

\usepackage{graphicx}
\graphicspath{{figures/}}
\usepackage{float}
\usepackage{caption}
\usepackage{subcaption}

\usepackage{threeparttable}
\usepackage{longtable}
\usepackage{multirow}
\usepackage{booktabs}
\usepackage{makecell}
\usepackage{pdflscape}

\usepackage{csquotes}
\usepackage{pifont}
\usepackage{xspace}
\usepackage{amsmath,amssymb}
\usepackage{enumitem}

\usepackage{hyperref}
\usepackage{url}
\usepackage{cleveref}

\newcommand{\code}[1]{\texttt{#1}\xspace}
\newcommand{\pkg}[1]{\textit{#1}\xspace}
\newcommand{\Rstats}{\textsf{R}\xspace}
\newcommand{\see}[2][]{(\Cref{#2}#1)\xspace}
\newcommand{\etal}{\textit{et al.}\xspace}

\title{A Large-Scale Neutral Comparison Study of Survival Models on Low-Dimensional Data}

\author[1,2,3,4]{Lukas Burk\thanks{Corresponding author. \href{mailto:burk@leibniz-bips.de}{burk@leibniz-bips.de}}}
\author[5,8]{John Zobolas}
\author[1,2]{Bernd Bischl}
\author[1,2]{Andreas Bender}
\author[3,4]{Marvin N. Wright}
\author[6]{Raphael Sonabend}

\affil[1]{Department of Statistics, LMU Munich, Munich, Germany}
\affil[2]{Munich Center for Machine Learning, Munich, Germany}
\affil[3]{Leibniz Institute for Prevention Research and Epidemiology -- BIPS, Bremen, Germany}
\affil[4]{Faculty of Mathematics \& Computer Science, University of Bremen, Bremen, Germany}
\affil[5]{Department of Cancer Genetics, Institute for Cancer Research, Oslo University Hospital, Oslo, Norway}
\affil[6]{OSPO Now, London, UK}
\affil[8]{Department of Biostatistics, Oslo Centre for Biostatistics and Epidemiology (OCBE), University of Oslo (UiO), Oslo, Norway}

\date{}

\begin{document}

\maketitle

\begin{abstract}
\noindent\textbf{Motivation:}
This work presents the first large-scale neutral benchmark experiment focused on single-event, right-censored, low-dimensional survival data.
Benchmark experiments are essential in methodological research to scientifically compare new and existing model classes through proper empirical evaluation.
Existing benchmarks in the survival literature are smaller in scale regarding the number of used datasets and extent of empirical evaluation.
They often lack appropriate tuning or evaluation procedures, while other comparison studies focus on qualitative reviews rather than quantitative comparisons.
This comprehensive study aims to fill the gap by neutrally evaluating a broad range of methods and providing generalizable guidelines for practitioners.\\[4pt]
\noindent\textbf{Results:}
We benchmark 19 models, ranging from classical statistical approaches to many common machine learning methods, on 34 publicly available datasets.
The benchmark tunes models using both a discrimination measure (Harrell's C-index) and a scoring rule (Integrated Survival Brier Score), and evaluates them across six metrics covering discrimination, calibration, and overall predictive performance.
Despite superior average ranks in overall predictive performance from individual learners like oblique random survival forests and likelihood-based boosting, and better discrimination rankings from multiple boosting- and tree-based methods as well as parametric survival models, no method significantly outperforms the commonly used Cox proportional hazards model for either tuning measure.\\[4pt]
\noindent We conclude that for predictive purposes in the standard survival analysis setting of low-dimensional, right-censored data, the Cox Proportional Hazards model remains a simple and robust method, sufficient for most practitioners.\\[4pt]
\noindent\textbf{Availability and Implementation:} All code, data, and results are publicly available on GitHub \url{https://github.com/slds-lmu/paper_2023_survival_benchmark}\\[4pt]
\noindent\textbf{Contact:} \url{burk@leibniz-bips.de}
\end{abstract}


\section{Introduction}
\label{sec:intro}

Survival analysis is an important branch of statistics for data where the outcome is the time until an event of interest occurs.
Such data often exhibits incomplete information about the outcome, for example due to censoring.
Traditionally applied in medical research to estimate how patient survival relates to clinical or biological characteristics (features), it has been applied in a broad range of applications across various domains.
By effectively incorporating information from both completed and ongoing (censored) cases, survival analysis can yield accurate and informative predictions.
This capability is invaluable in fields such as medicine, finance, and in different industrial sectors, where risk prediction under censoring is an important decision making component.
Many methods have been introduced in this field, from the Cox Proportional Hazards (CPH) \citep{cox1975, cox1972} and Accelerated Failure Time (AFT) models \citep{kalbfleisch2011} to tree-based methods including Random Survival Forests (RSFs) \citep{ishwaran2008} and Gradient Boosting Machines (GBMs) \citep{friedman1999,chen2016xgboost,barnwal2022} as well as many others.
Throughout this paper, we consider only the right-censored, single-event survival setting, which is very common in practical applications.
More complex settings with competing risks and different types of censoring and truncation are becoming more relevant, but are currently not broadly supported by machine learning algorithms, particularly their respective implementations.
In a nutshell, right-censoring occurs when some subjects did not experience the event of interest, either due to drop-out or the end of the study (both assumed to be unrelated to the event of interest here).
Formally, let $Y_i\sim F_Y; Y_i> 0; i=1,\ldots,n$ be random variables representing the event times and $C_i\sim F_C;\ C_i>0$ the censoring time.
In right-censored data, we do not observe realizations of $Y_i$ but rather of the tuple $(T_i = \min(Y_i, C_i), D_i = \mathbb{I}(Y_i \leq C_i))$.
The goal of survival analysis is to estimate the distribution $F_Y$, or quantities derived from it, e.g., the expected survival time $\mathbb{E}_Y(Y)$, based on realizations of $(T_i, D_i)$ and features $\mathbf{x}_i$; see Section \ref{sec:methods} for definition of prediction types.
The observed data is given by tuples $(t_i, d_i, \mathbf{x}_i),\ i=1,\ldots,n$, where $t_i$ is the \emph{observed} outcome time (either event or censoring time, whichever occurred first), $d_i$ is the status indicator (0 if observation is censored, 1 if the event of interest was observed) and $\mathbf{x}_i$ is the feature vector.

\textbf{Contributions}:
This paper introduces the first neutral, large-scale comparison study for single-event, right-censored survival data with a large number of datasets (34), models (19), tuning measures (2), and evaluation measures (6) included.
We benchmark survival techniques in the low-dimensional setting, which represents a type of data that practitioners often encounter.
The scale of our study, encompassing diverse datasets (see Table \ref{tab:datasets}) and an extensive evaluation procedure using multiple metrics, provides a robust benchmark for comparing model performance in right-censored, low-dimensional settings~\citep{herrmann2020}.
Furthermore, we run a \enquote{neutral} benchmark in accordance with the guidelines laid out by \cite{boulesteix2013} which we further outline in \Cref{sec:exp_design}.
Based on our review of the literature (see \Cref{sec:lit}), there is no other study: a) with a comparable number of datasets or methods; b) that compared methods after sufficient tuning for both discrimination and for overall predictive ability as measured by the integrated survival Brier score (ISBS); or c) that neutrally compares methods.
Finally, our experiment code is available on GitHub and our collection of datasets is available as an OpenML benchmark suite~\citep{openml2013} and our hyperparameter search spaces will be available in a forthcoming release of \pkg{mlr3tuningspaces}~\citep{pkgmlr3tuningspaces}.

\section{Literature Review}
\label{sec:lit}

The experiments described in this paper provide a comparison of both classical and machine learning (ML) survival models in a low-dimensional setting.
We use these broad terms for model classes in analogy to the taxonomy provided by \cite{wang2019}, where the term \enquote{classical} refers to semi- and fully parametric methods such as the CPH and AFT models or derivatives thereof, including penalized variants.
\enquote{ML} here refers to non-linear and non-parametric methods ranging from tree-based methods including RSFs, boosting approaches such as GBMs or likelihood-based boosting (CoxBoost), to Survival Support Vector Machines (SSVMs), Artificial Neural Networks (ANNs) and Deep Learning (DL) methods.
Although there is no objective and generally accepted definition to determine whether a dataset is \enquote{high-dimensional}, we colloquially define it to refer to scenarios where the number of features exceeds the number of observations ($p >> n$).

Historically, surveys, reviews, and analytical comparisons of survival models can be grouped into:
1) empirical comparisons of models with limited scope;
2) qualitative surveys without benchmark experiments.
We provide a short overview of the literature for comparisons of survival models, with an extended review available in Appendix \ref{app:lit}.
Papers that empirically compare survival models (1) are further separated into studies that:
(i) compare `classical' models only; (ii) compare multiple ML and classical model classes; (iii) compare one novel model (or class) to one or more baseline models; (iv) exclusively focus on high-dimensional data.

\textbf{Comparisons of Classical Models} often compare CPH and AFT models, including \citep{moghimi-dehkordi2008,georgousopoulou2015,zare2015,habibi2018} showing both methods yielding similar hazard ratios but without evaluation metrics on independent test data, relying on graphical procedures to draw conclusions; \citep{dirick2017} additionally compare flexible Cox models including splines using time-dependent area under the Receiver Operating Characteristic curve (AUC), mean squared error, and mean absolute error, finding that the CPH with penalized splines outperform the other models.

\textbf{Comparisons of ML and Classical Models}
The experiments carried out in this paper belong in this category.
Only two prior experiments could be found that neutrally benchmarked more than one ML model class on low-dimensional data.
\cite{kattan2003} benchmarked tree-based models, ANNs and CPH with Harrell's C-index across three datasets.
The models are compared for significant differences using 50 times repeated nested cross-validation, but the authors do not clarify their tuning procedure.
Boxplots across all replications indicate that no machine learning model outperformed the CPH.
The authors note the small number of datasets used for comparison as their primary limitation.
\cite{zhang2021survbenchmark} compare classical and ML methods, taking into account feasibility and computational efficiency for various tasks in the biomedical field.
Methods are evaluated on six clinical and 16 omics datasets using 11 metrics, including time-dependent AUC, ISBS and multiple variations of the C-index.
Critically, methods were applied with implementation-specific default hyperparameter settings without tuning, severely limiting the generalizability of their results.

\textbf{Comparisons of a Novel Model Class} include~\cite{luxhoj1997, ohno-machado1997} benchmarking newly developed ANNs against CPH and \cite{goli2016a} comparing modern Support Vector Regression variants against CPH, where neither found significant differences.
\cite{jaeger2024} compare a novel implementation of oblique RSFs (ORSF) to the previous implementation, as well as other RSFs, GBMs, penalized CPH, and ANNs.
Using Harrell's C and ISBS, they find that ORSFs outperform GBMs and penalized CPH, but with only minimal tuning applied for some models.

\textbf{Comparisons on High-Dimensional Data} have gained popularity in the survival literature, with many recent studies focusing on the area of multi-omics data~\citep{zhao2024tutorialsurvival}.
\cite{herrmann2020} perform a large-scale benchmark including penalized regression, GBMs, and RSFs, evaluating on Harrell's C and ISBS without finding any significant differences between the different model classes .
\cite{spooner2020} compare a similar group of models evaluating on Harrell's C only, finding few significant differences.
\cite{wissel2023systematic} compare DL methods, RSFs and CPH, evaluating on Antolini's C-index and ISBS with a focus on noise-resistance, noting a lack thereof for all models.

\section{Benchmark Experiments}
\label{sec:exp}

\subsection{Study Design}
\label{sec:exp_design}
The experiments in this study are designed to assess the status quo of survival models, including both classical and machine learning approaches.
In order to achieve this objective, this study aims to be a \enquote{neutral comparison study}~\citep{boulesteix2013}.
Following the guidelines put forward by Boulesteix~\etal, such studies:

\textbf{Focus on model comparison}: The focus of this study is on model comparison rather than on the examination of a novel model.
We do not favor one dataset over another and draw conclusions across all datasets instead of trying to find data sets in which the models performed well.

\textbf{Fair and neutral setup}: At least one representative of all methods compared in this experiment was contacted, and hyperparameter configurations were discussed with all who responded.
Every maintainer of the used software packages implementing the evaluated methods was given equal opportunity to influence the experiment, ensuring there was no bias in model configuration.
We are grateful for the maintainers' time supporting this effort.

\textbf{Model, performance measures, and data are chosen in a rational way}:
The study is designed to assess the status quo, which excludes models and measures that have been published without peer-review.
We focus on a single primary metric to evaluate discrimination and another for overall predictive performance, while additional measures are reported for comparison purposes.
To align with common practice and to allow for general comparability, we also assess models by measures even if these are known to be flawed, e.g., increasing bias of Harrell's C-index for increasing censoring percentages (see, for example, \cite{schmid2012}).
The inclusion criteria for datasets were as follows:
We use real-world datasets that include at least two features, a right-censoring indicator with a survival time, have at least 100 observed events, and do not qualify as high-dimensional, i.e., have fewer features than observations.
We exclude datasets with competing risk endpoints, recurrent events, or other non-standard settings such as left-censoring.
No quota was specified regarding censoring proportions in the datasets.

\textbf{Implementation, Reproducibility, and Accessibility}
Experiments were conducted on \Rstats 4.4.3 on the Beartooth Computing Environment \citep{beartooth}. 
All code required to run the experiments and generate the results is available in a public \href{https://github.com/slds-lmu/paper_2023_survival_benchmark}{GitHub repository}\footnote{\url{https://github.com/slds-lmu/paper_2023_survival_benchmark}} licensed under GPL-3.
Further details on software used are available in Appendix \ref{app:reprodubility}.
For additional reproducibility, our hyperparameter search spaces will be published with an upcoming release of \pkg{mlr3tuningspaces} \citep{pkgmlr3tuningspaces} and our datasets will be available as an OpenML benchmark suite \citep{openml2013} while already being available from GitHub.

\subsection{Models and Configurations}\label{sec:exp_mod}

The algorithms compared in this experiment were chosen by identifying commonly used models with readily available implementations:
\textbf{(1)} Kaplan-Meier (KM)~\citep{kaplanmeier1958};
\textbf{(2)} Nelson-Aalen (NEL)~\citep{aalen1978};
\textbf{(3)} Akritas Estimator (AK)~\citep{akritas1994};
\textbf{(4)} Cox PH (CPH)~\citep{cox1975};
\textbf{(5)} Regularized CPH with cross-validation (GLMN)~\citep{simon2011};
\textbf{(6)} Penalized (Pen)~\citep{goeman2010l1penalized};
\textbf{(7)} Parametric AFT (AFT)~\citep{kalbfleisch2011};
\textbf{(8)} Flexible Splines (Flex)~\citep{roystonparmar2002};
\textbf{(9)} Random Survival Forest (RFSRC)~\citep{ishwaran2008};
\textbf{(10)} Random Survival Forest (RAN)~\citep{pkgranger};
\textbf{(11)} Conditional Inference Forest (CIF)~\citep{hothorn2006};
\textbf{(12)} Oblique Random Survival Forest (ORSF)~\citep{jaeger2024};
\textbf{(13)} Relative Risk Tree (RRT)~\citep{breiman1984};
\textbf{(14)} Model-Based Boosting with Cox and AFT objective (MBSTCox and MBSTAFT)~\citep{buhlmann2003boostingl2};
\textbf{(15)} CoxBoost (CoxB)~\citep{binder2008allowing};
\textbf{(16)} XGBoost with Cox and AFT objective (XGBCox and XGBAFT)~\citep{chen2016xgboost,barnwal2022};
\textbf{(17)} Survival-SVM (SSVM)~\citep{vanbelle2011supportvector}.

The full table of all models, including respective software packages, is given in Appendix \ref{app:models}, and the online supplement includes recorded versions of all packages involved in the experiment.
In our selection, we focused on well-established models with generally robust implementations, provided as well-maintained packages or wrapper functions within benchmarking software.
This excludes some recently proposed DL based methods like DeepSurv~\citep{pkgdeepsurv} and DeepHit~\citep{pkgdeephit}, which have higher computational complexity, require intensive tuning, and in initial experiments could not be evaluated reliably within our benchmark suite due to a lack of stable implementations.
The KM and NEL estimators are used as non-parametric baselines, while AK acts as a more flexible baseline as it estimates a conditional survival function without relying on independent or non-informative censoring.
An additional table in Appendix \ref{app:models} lists the associated hyperparameter and pre-processing configurations.
Note that while some studied methods could be considered algorithmically equivalent, e.g., RAN and RFSRC both implement Random Survival Forests, their implementations are not, and for that reason, we opted to include such variants in the comparison.
A small number of methods provide internal optimization methods for certain hyperparameters, which we used when available / possible in our global tuning procedure.
This affects both XGBoost models and their early-stopping mechanism to tune the \code{nrounds} parameter and GLMN for \code{lambda}, while CoxB is exclusively tuned using its internal mechanism.
For further details we refer to the experiment code on GitHub.

\subsection{Resampling, Tuning, Prediction Types, and Pre-Processing}
\label{sec:methods}

\textbf{Resampling} is performed as nested repeated cross-validation with three outer and three inner resampling folds for unbiased generalization error estimates~\citep{bischl2023hpo}.
Outer cross-validation is repeated five times for datasets with a total number of events in $[500, 2500)$ and ten times for datasets with fewer than 500 events.
This aides the stability of performance estimates by increasing the total number of evaluations to up to 30 (10 repetitions $\times$ 3 outer folds) for the datasets with few events, while saving computational costs for the largest tasks where the available training data is likely sufficient for reliable evaluation.
Inner cross-validation is always repeated twice, leading to six total evaluations per tuning iteration.
Resampling is stratified by event indicator to preserve the censoring proportion in respective folds.

\textbf{Tuning} is performed on the inner resampling folds of the nested cross-validation.
We use Bayesian optimization~\citep{garnett2023bayesoptbook, bischl2023hpo} with $50\rho$ iterations, where $\rho$ is the dimensionality of the search space, ranging from 1 to 8.
For example, one parameter of GLMN is tuned for 50 iterations, whereas six parameters of ORSF are tuned for 300 iterations.
This method grants each method the opportunity to be tuned to the same relative amount: 50 iterations per tunable parameter.
When the tuning space was finite and smaller than 50, we exhausted all configurations to achieve the same tuning result but with lower computational cost; this applied to learners AFT (tuning across three distribution families), Flex (tuning $k \in \{2, \ldots, 10\}$, and RRT (tuning $\mathtt{minbucket} \in \{5, 6, \ldots, 50\}$).
We set a time limit for the tuning process of 150 hours to ensure that one outer resampling iteration (tuning and final model fitting) could be completed within seven days, a constraint imposed by the computational environment.
This restriction was most frequently violated for memory-intensive models on datasets with many observations, where in some cases the final evaluation was unsuccessful (see \Cref{sec:results} and Appendix \ref{app:errors}).
The tuning process is repeated independently for each tuning measure (see \Cref{sec:performance-eval}).

\textbf{Prediction types:}
In general, there are four prediction types in survival analysis \citep{sonabend2021b}:
A linear predictor \code{lp}, continuous ranking \code{crank} (e.g., a relative risk), a distribution \code{distr} (e.g., the survival probability), and predicted survival times \code{response}.
The \code{response} time is very uncommon due to its generally poor quality \citep{henderson2001accuracypoint} and only directly provided by the Survival SVM and XGBAFT at the time of writing.
We focus on evaluating distribution and continuous rank predictions.
The prediction types provided by individual methods (and implementations) can vary, which is why \pkg{mlr3proba} \citep{pkgmlr3proba} compositors are used to derive missing prediction types needed for evaluation when appropriate.
Where models only predict a probabilistic prediction, \code{crank} is calculated as the expected mortality derived from the \code{distr} prediction~\citep{ishwaran2008,sonabend2022chack}.
When models predict only \code{crank} or \code{lp}, then we exclusively evaluate them on discrimination measures, which is the case for RRT, XGBAFT, MBSTAFT, and SSVM.
The XGBCox predicts \code{lp}, and the \code{distr} prediction is composed using the Breslow estimator~\citep{lin2007} similarly to previous benchmarks~\citep{jaeger2024}.

\textbf{Pre-processing} is applied only if either technically required to run a model or in line with standard recommendations for that model class.
This includes standardization of covariates to unit variance and zero mean and/or dummy encoding of categorical features.
We created pipelines for all learners (where required) with \pkg{mlr3pipelines} \citep{pkgmlr3pipelines}, to combine the respective pre-processing operation with the learning algorithm, and to properly embed the pre-processing into the cross-validation.
Appendix \ref{app:models} lists the model-specific pre-processing performed.
In addition to these model-specific pre-processing steps, we collapse levels of categorical variables with frequencies below 5\% as part of the model pipeline, ensuring that high-cardinality categorical features are handled consistently.
As we only applied basic pre-processing, no additional hyperparameters were added to the tuning search space.

\subsection{Performance Evaluation}\label{sec:performance-eval}

We assess performance using two primary measures alongside four additional measures.
For cases where individual model predictions were not possible during the inner- or outer resampling procedure due to any kind of error (computational issues, non-convergence, etc.), the prediction of the KM estimator was used as a fallback result.
This ensures a statistically sound evaluation, and is considered a reasonable compromise between either overpenalizing models by inserting some constant value which would depend on the given metric and dataset, or simply disregarding failed iterations during evaluation, which is overly optimistic~\citep[see e.g.][]{mlr3booklargescale}.

\textbf{Measures} chosen for this benchmark are summarized in \Cref{tab:measures}.
Of these measures, only two are used to provide primary results: Harrell's C \citep{harrell1982} for pure discrimination and the Integrated Survival Brier Score (ISBS)~\citep{graf1999} for overall predictive ability, including calibration.
The benchmark procedure is run twice, tuning either for discrimination (Harrell's C) or overall predictive ability (ISBS) and evaluated according to similar measures.
For integrated measures, we use the commonly chosen 80\% quantile as the upper bound \citep{sonabend2025examiningmarginal}.
We additionally explore the calibration measures D-Calibration \citep{haider2020effectivewaysa} and van Houwelingen's $\alpha$ \citep{vanhouwelingen2000}, which we only apply on models tuned with ISBS.

\textbf{Statistical Analysis} is conducted following Dem\v{s}ar~\citep{demsar2006}, initially performing global Friedman rank sum tests for all measures, where the \enquote{groups} are the models and the \enquote{blocks} are the independent datasets.
Significance after Bonferroni-Holm adjustment determines whether post-hoc tests are conducted.
Post-hoc (multiple-testing corrected) Bonferroni-Dunn tests are conducted and presented as critical difference diagrams, using CPH as the reference model for comparison.

\begin{table}[h]
    \caption{Performance measures used in the benchmark, including their type and the quantity they evaluate, i.e., continuous ranks (crank) or a survival distribution (distr). Harrell's C and ISBS are used for primary analysis with remaining results in Appendix \ref{app:results}. Models evaluated with each measure were tuned on the corresponding tuning measure.}
    \label{tab:measures}
    \centering
    \small
    \begin{tabular}{@{}llll@{}}
        \toprule
        \textbf{Measure} & \textbf{Tuned on} & \textbf{Type} & \textbf{Evaluates} \\
        \midrule
        Harrell's C & Harrell's C & Discrimination & \code{crank} \\
        Uno's C & Harrell's C & Discrimination & \code{crank} \\
        ISBS$^a$ & ISBS & Scoring Rule & \code{distr} \\
        ISLL$^b$ & ISBS & Scoring Rule & \code{distr} \\
        D-Calibration & ISBS & Calibration & \code{distr} \\
        $\alpha$-Calibration$^c$ & ISBS & Calibration & \code{distr} \\
        \bottomrule
    \end{tabular}
    \\[2pt]
    \footnotesize
    $^a$Integrated Survival Brier Score
    $^b$Integrated Survival Log-Likelihood
    $^c$van Houwelingen's $\alpha$
\end{table}

\subsection{Datasets}\label{sec:data}

To obtain a collection of suitable datasets, we ran a search across the CRAN Task View \enquote{Survival Analysis}\footnote{\url{https://cran.r-project.org/view=Survival}}, Python's \pkg{pycox} library and related literature (see \Cref{sec:lit}) and existing collection of survival datasets~\citep{drysdale2022survsetopensource}, yielding over 120 datasets.
After applying the dataset inclusion criteria (see \Cref{sec:exp_design}) and removing duplicates and derivations of other datasets, a total of 34 datasets remained.
Minor changes are made to variable names, recoding of factor levels, and deletion of non-informative or ``illegal'' covariates like ID numbers.
Observations are deleted if their event time is equal to zero.
Since this benchmark is not concerned with a model's ability to handle or impute missing covariate data, observations with missing values were removed, which occurred very rarely.
Lastly, in cases where datasets had a large number of unique time points, the time variable was coarsened by appropriate rounding, greatly reducing computational cost for some methods, including RAN, AK, and CIF.
For full details, see the pre-processing code contained in the GitHub repository (see \Cref{sec:exp_design}).
Summaries of the datasets in terms of the number of observations and covariates after modification and censoring proportions, along with citations for the respective sources, can be found in Appendix \ref{app:datasets}.

\section{Results}
\label{sec:results}

Global Friedman tests were significant for all measures, indicating the presence of significant differences between models and allowing for post-hoc analysis.
The number of times models failed to compute results due to either time or memory constraints and required the score imputation using KM \see{sec:methods} is tabulated in Appendix \ref{app:errors}.
We present critical difference (CD) plots for the baseline-comparison to CPH for both discrimination and overall performance (see \Cref{sec:performance-eval}, \cite{demsar2006}).
The top line of a CD plot represents a model's average performance rank across all datasets in the benchmark, where a lower ranking implies better performance regardless of the evaluation measure applied.
Thick horizontal lines around the CPH model rank indicate the symmetric critical difference, meaning that other models within this range do not significantly differ in rank from the reference model.

\subsection{Discrimination}
CD plots for discrimination, tuned and evaluated on Harrell's C (\Cref{fig:cd-bd}, top), indicate that all models outperform the baseline learners (KM, NA, AK) as expected.
MBSTAFT, AFT, RAN, and CoxB are the top-performing models but fail to significantly outperform the CPH baseline.
The remaining models mostly belong to the classes of RSFs and GBMs, which all achieve average ranks between 6.5 and 8, indicating similar discrimination performance.
Pen, XGBCox, XGBAFT, and Flex fall slightly behind, but still lie within the critical difference.

\subsection{Overall Performance}
\begin{figure}[htbp]
\centering
\includegraphics[width=\textwidth]{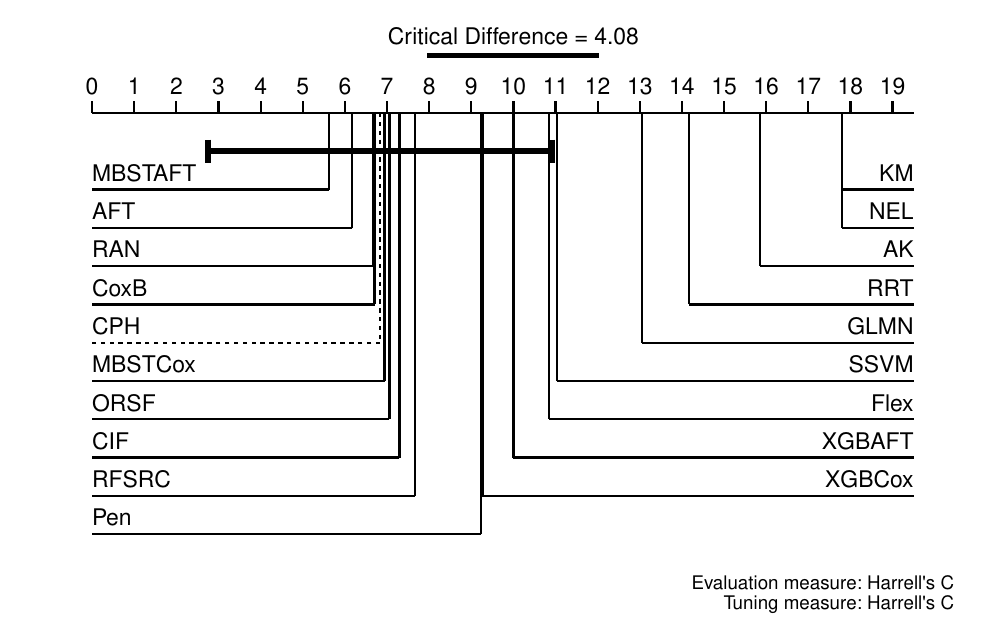}
\\[8pt]
\includegraphics[width=\textwidth]{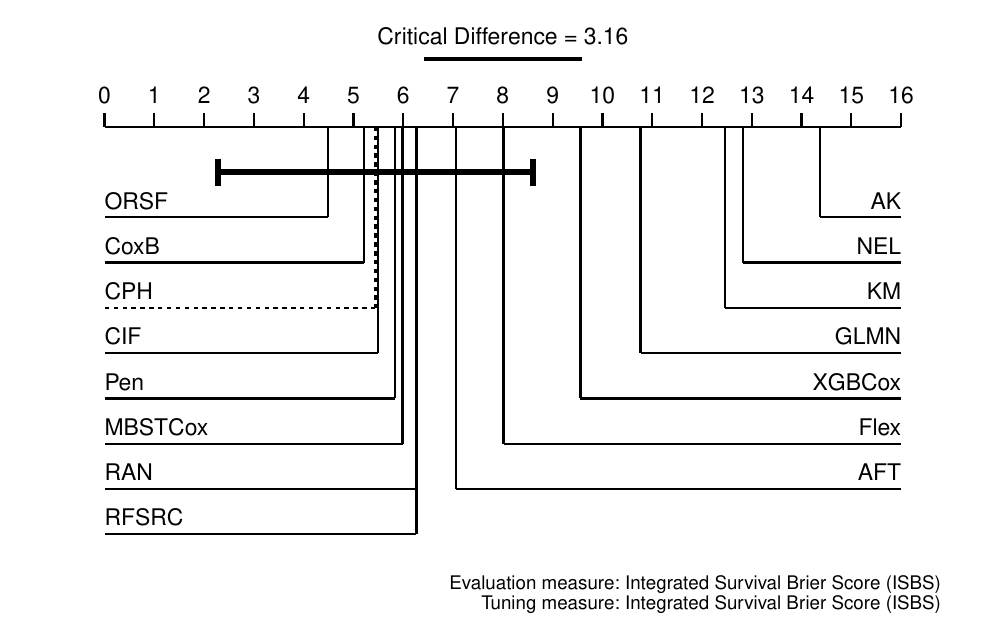}
\caption{Critical difference plots comparing models with the CPH reference tuned and evaluated on either Harrell's C (top) or ISBS (bottom). Superior models (lower ranking scores) are on the left with decreasing performance (higher rank) moving right. Models connected by thick horizontal lines are not significantly different from the baseline when adjusting for multiple comparisons.}
\label{fig:cd-bd}
\end{figure}

The CD plot for overall performance tuned and evaluated on ISBS (\Cref{fig:cd-bd}, bottom) similarly indicates that no model significantly outperforms CPH. Only
ORSF and CoxB perform better, while CIF ranks almost identically to CPH.
Pen, MBSTCox, RAN, and RFSRC fall slightly behind, with AFT and Flex further behind.
XGBCox and GLMN are the only non-baseline models significantly outperformed by CPH.

\begin{figure}[htbp]
    \centering
    \includegraphics[width=0.9\textwidth]{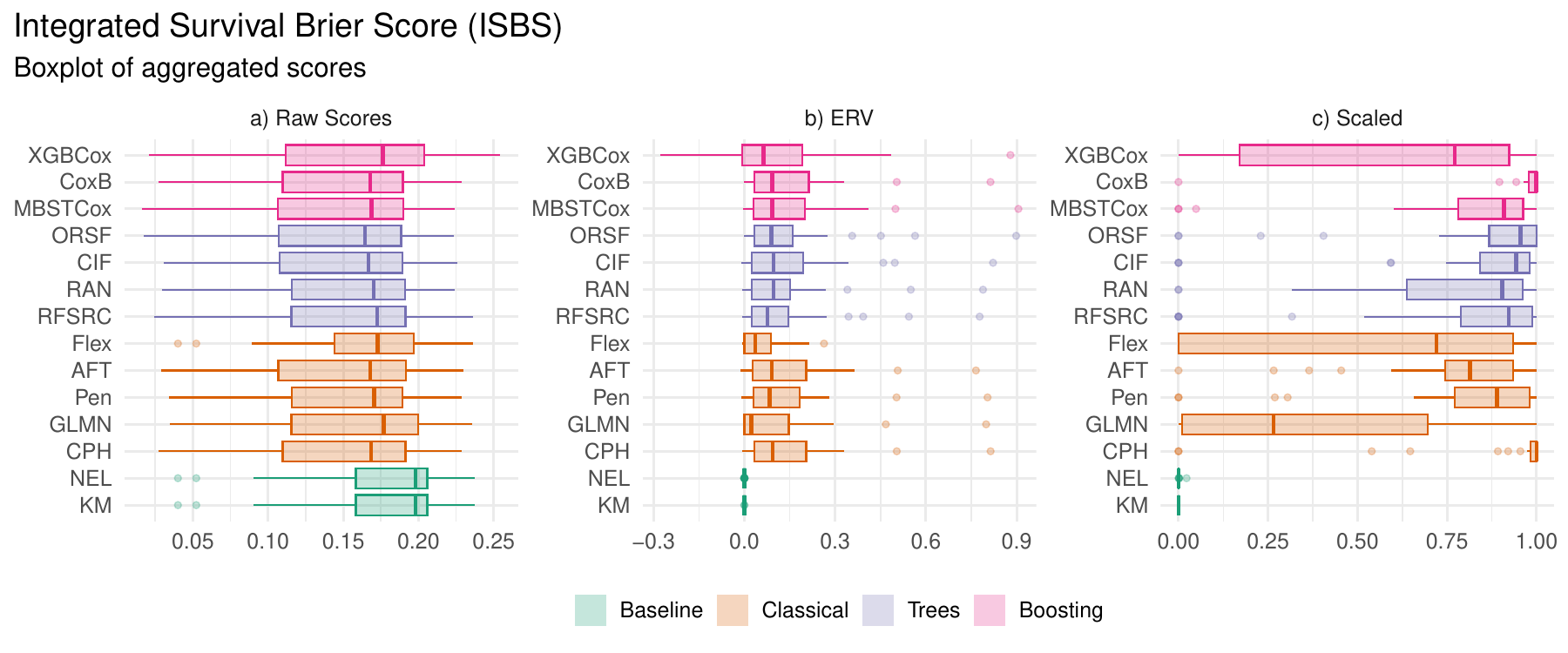}
    \caption{Boxplots of aggregated scores across all datasets for models tuned and evaluated with ISBS showing unmodified ISBS scores (a), Explained Residual Variation (ERV) scores (b), and scores scaled such that 0 is equivalent to KM and 1 is achieved by the best model for each dataset and measure. AK is omitted due to largely off-the-scale negative ERV values to avoid scaling issues.}
    \label{fig:boxp-scalings}
\end{figure}

We additionally present boxplots both for individual scores per dataset and aggregated scores.
We offer three versions of these aggregated scores to support evaluation and analysis, illustrated by \Cref{fig:boxp-scalings}:
a) Raw scores as calculated by the corresponding measure; %
b) \enquote{Explained Residual Variation} (ERV)~\citep{korn1991} scores similar to the \enquote{Index of Prediction Accuracy}~\citep{kattan2018indexprediction} where negative values imply performance worse than KM, 0 is equivalent to KM, and 1 denotes a perfect model;
c) Scaled scores, whose interpretation is the same as the ERV ones, with the difference that 1 is achieved by the best model for a given task and measure \citep{caruana2006empiricalcomparison}.

\subsection{Calibration}

Calibration results presented in this section, measured by D-Calibration and van Houwelingen's $\alpha$ (see Table \ref{tab:measures}), were obtained by tuning all considered model for ISBS.
We hope that results for these measures provide additional insights, but would like to point out that they might be considered somewhat experimental with respect to underlying theory and respective implementation.

\textbf{D-Calibration} considers a model well-calibrated if the p-value of the underlying test is greater than 0.05, which we display in the form of a heatmap with X indicating a significant test result, indicating a model is not well-calibrated.
Here, NEL, RFSRC, ORSF, and RAN are among the best calibrated, while AK, XGBCox, CIF, AFT and CPH would be considered poorly calibrated.

\begin{figure}[ht]
    \centering
    \includegraphics[width=\textwidth]{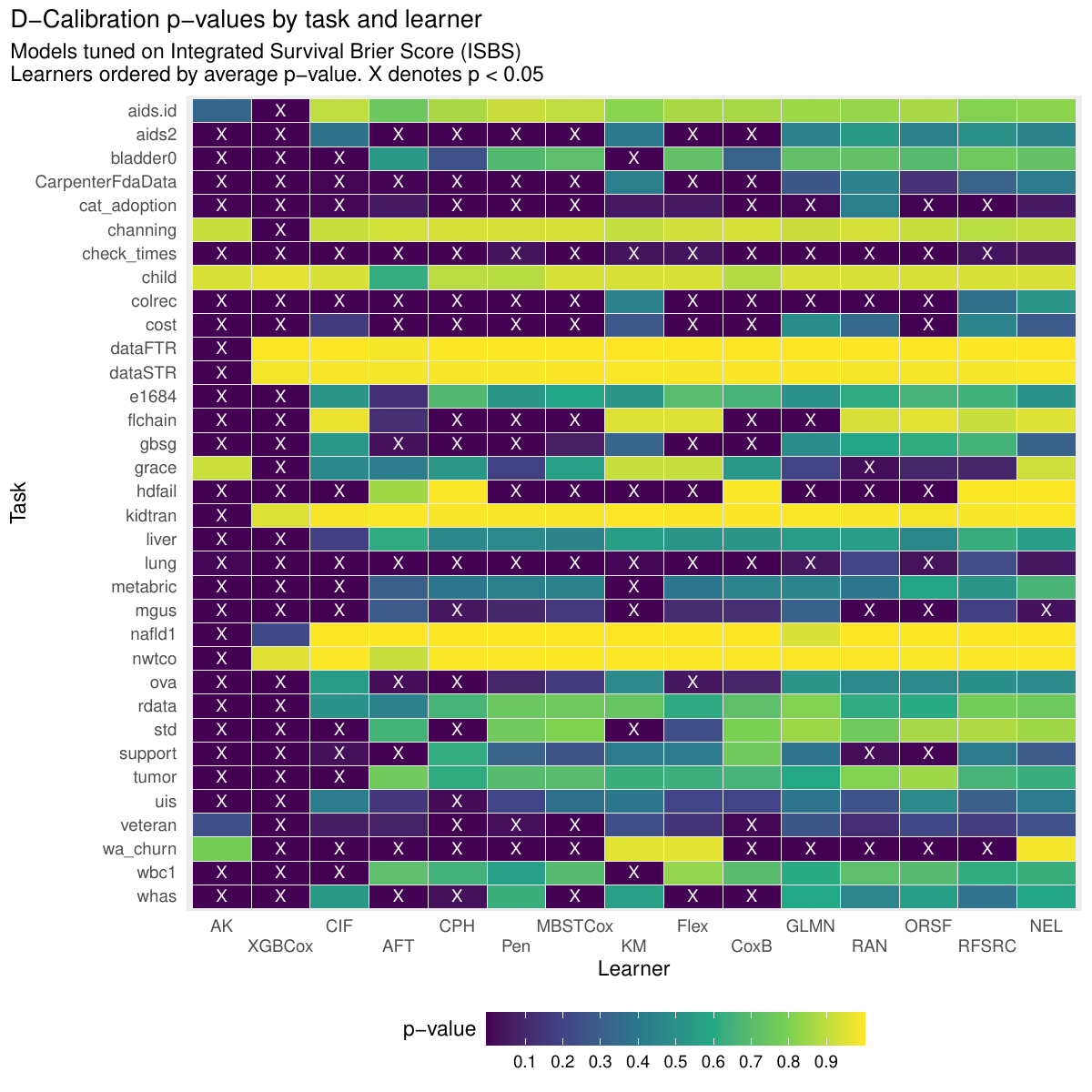}
    \caption{D-Calibration p-value heatmap across all datasets and models. `X' indicates $p < 0.05$ while a non-significant result indicates good calibration.}
\end{figure}

\textbf{Van Houwelingen's} $\alpha$ relates predicted and observed hazards, with a value close to 1 implying a well-calibrated model.
Here, KM, NEL, GLMN, and Pen show good calibration, while XGBcox shows a wide spread indicating poor calibration.
RFSRC, RAN, and ORSF similarly show more skewed scores, indicating poor calibration.

\begin{figure}[ht]
    \centering
    \includegraphics[width=\textwidth]{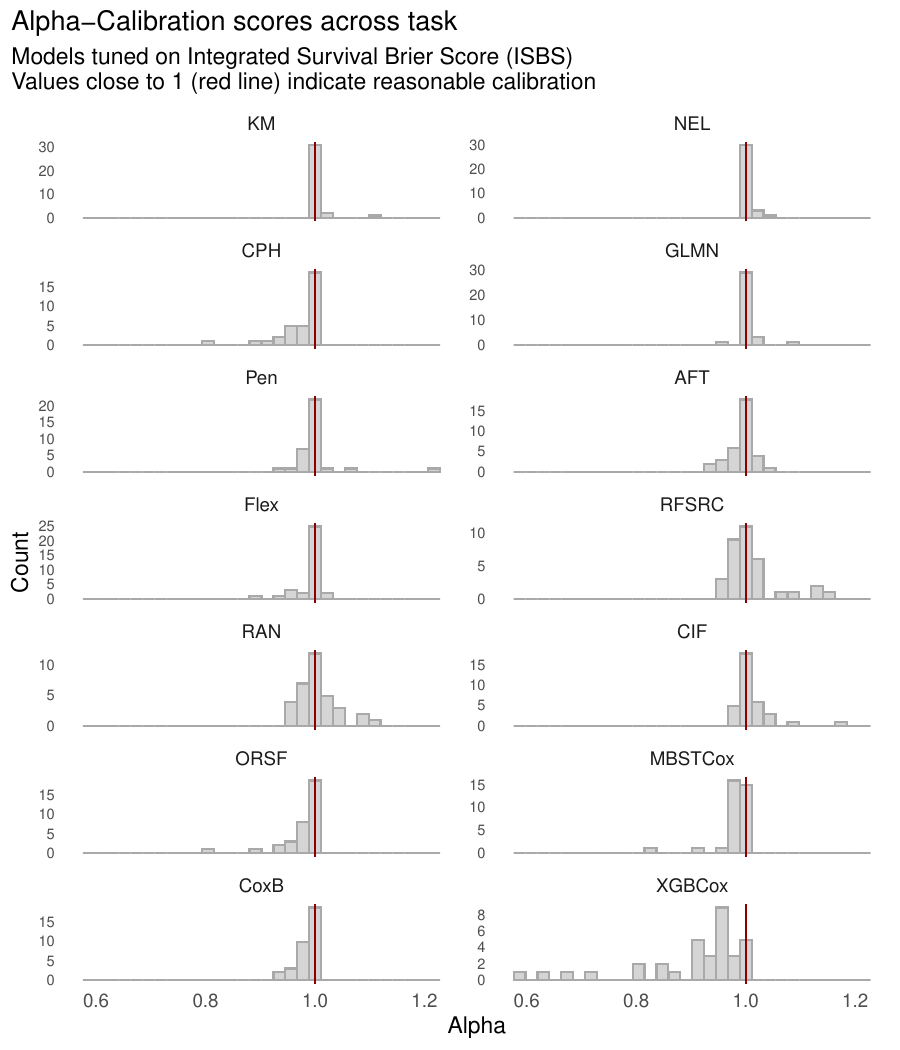}
    \caption{Calibration scores across all datasets for each model using van Houwelingen's $\alpha$ indicate good calibration when values are close to 1. }
\end{figure}

Further results for all measures can be found in Appendix \ref{app:results}.
Additional tables and visualizations are available on our results website which is linked in our GitHub repository.
The repository also provides downloads for all raw results, including tuning archives for the close to 20.000 individual tuning procedures performed in this experiment.

\section{Discussion}
\label{sec:discussion}

\textbf{Discrimination}
The critical difference plots show that MBSTAFT, AFT, RAN, CoxB, and CPH were the five highest-scoring methods, notably including two classical methods.
The best-ranking ML methods both used the AFT model form but do not significantly outperform CPH.
XGBoost with either Cox or AFT objective appears at the lower end of the models' ranking, on par with CPH, while ranking lower than the likelihood-boosting method CoxB and all RSFs.
Conversely, CoxB did not require any explicit tuning outside of its internal optimization method, making it computationally more efficient while achieving comparable or better discrimination performance.
Penalized Cox (GLMN), decision trees (RRT) and the SSVM are the only non-baseline models outperformed by CPH, which in either case may be expected given that RRT is known to be inferior to boosted or bagged trees for prediction purposes, and GLMN represents a penalized version of CPH that may not be well suited for the low-dimensional tasks included in this benchmark, while Pen, on the other hand, is a similar method that did rank better by some margin.
AK performs as expected, representing a slight improvement over KM and NA.
KM and NA are included solely as a reference because their Harrell's C scores are intentionally constant at 0.5.
We note that our use of the parametric AFT model can be considered unconventional, as we tune the distribution family (i.e., whether to use a Weibull, log-normal, or log-logistic distribution) within the tuning iterations, thus performance evaluation in the outer resampling iterations could be based on a different distribution.
This also affects XGBAFT analogously.
In real-world scenarios, a specific distribution is typically chosen in advance.

\textbf{Overall Performance}
Overall performance is judged by the ISBS scoring rule~\citep{sonabend2025examiningmarginal} and should capture both a model's discrimination and calibration ability.
We can only evaluate models on ISBS which provide a survival distribution prediction, therefore omitting  RRT, XGBAFT, MBSTAFT, and SSVM in this setting as described in Section \ref{sec:methods}.

CoxB again performs quite well, indicating it is an effective method in both settings.
CPH in third place shows its general effectiveness, being on par with the computationally more complex CIF.
Pen again outperforms GLMN by a wide margin despite their similarity, and analogously, the GBMs with Cox-objective MBSTCox and XGBCox are also separated by more than three places in ranking.
The main difference between GLMN and Pen is that the former performs internal cross-validation for its primary regularization parameter, which in this setting could present a disadvantage.
For XGBCox, it should be noted that this method encountered many computational errors, leading to imputed performance values as described in Section \ref{sec:performance-eval}.
While the AFT-based approaches seemed to outrank the Cox-based ones on discrimination, here we see AFT falling behind CPH and other Cox-based methods.
The AK baseline shows noticeably worse performance than NEL and KM baselines, while it outperformed them in the discrimination setting.
Methods besides CPH that performed well using either evaluation metric include CoxB, ORSF, and MBSTCox.
Since not all methods could be evaluated on both metrics, the critical difference in each comparison is affected by the differing number of methods, complicating a direct comparison between the two resulting rankings.
We also note that many ML methods that perform on par with CPH on discrimination measures noticeably rank lower when considering overall performance measures like ISBS.

\textbf{Calibration}
In the case of D-Calibration, a model is considered well-calibrated if the underlying Pearson's $\chi^2$-test results in $p > 0.05$.
For van Houwelingen's $\alpha$, calibration is indicated by values close to 1.
All models, aside from notable outlier AK, as well as XGBCox and CIF, appear to be reasonably well calibrated based on D-Calibration, while results seem to depend largely on the dataset.
For van Houwelingen's $\alpha$, XGBCox is the least calibrated model apart from AK, which we omitted from the plot due to its greatly out-of-range scores.
This is consistent with the comparatively poor performance displayed in the results evaluating on ISBS, where both AK and XGBCox scored low.
We note that both calibration measures should be considered experimental, and more research on calibration-specific measures will be required for a more conclusive evaluation (e.g. \cite{austin2020graphicalcalibration})
Generally speaking, multiple measures should be considered for performance evaluation, as they may highlight aspects of performance relevant in different contexts.
For discrimination and overall performance, results on individual datasets are not discussed here but presented in Appendix \ref{app:results_per_dataset} as the overall trend is similar to the aggregated results, with expected variation between individual tasks.

\subsection{Limitations}
\label{sec:discussion_limit}

Our use of the AFT model prioritizes prediction over interpretation and could be simplified by splitting the tuned model into one model for each functional form (\enquote{Weibull}, \enquote{log-normal}, \enquote{log-logistic}), which would yield more interpretable results.
Since our focus lies on generalizability to low-dimensional, right-censored settings, our results will not generalize to more complex settings.
However, extension to left-censoring, competing risks, or other more complex endpoints first necessitates more comprehensive support by models and their implementations, as well as adaptation of available evaluation measures to these scenarios.
The number of datasets included could be extended, but still exceeds that of the vast majority of previous benchmarks.

\subsection{Conclusions and Future Work}
\label{sec:discussion_future}

Our results demonstrate that classical statistical methods, such as CPH and AFT, can in some settings significantly outperform complex ML algorithms in predictive survival tasks.
While it is possible to fine-tune and achieve better predictive accuracy using ML methods in individual cases (see Appendix \ref{app:results_per_dataset}), our results indicate that across a range of low-dimensional tasks, this is not the case in aggregate.
We therefore recommend practitioners to start with these conceptually and computationally simpler methods, and evaluate whether the additional computational cost and loss of interpretability are appropriate for their needs.
To improve upon our results, expanding this benchmark with a wider range of settings would be beneficial, allowing results to be generalized further, pending the corresponding software support.

\clearpage
\appendix
\section{Literature Review}\label{app:lit}

The following represents an extended literature review complementary to the summary provided in the main article.

\paragraph{Comparisons of ML and Classical Models}

The experiments carried out in this paper fall into this category.
Only two prior experiments could be found that neutrally benchmarked more than one ML model class on low-dimensional data.
%
\cite{kattan2003} benchmarked tree-based models, ANNs and CPH with Harrell's C-index across three datasets with varying censoring proportions.
The models are compared for significant differences by repeating the experiments up to 50 times with different seeds thus allowing for different hyper-parameter configurations and folds in cross-validation.
Boxplots across all replications indicate that no machine learning model outperformed the CPH.
%
\cite{zhang2021survbenchmark} compare classical and ML methods, taking into account feasibility and computational efficiency for various tasks in the biomedical field.
Methods are evaluated on six clinical and 16 omics datasets using 11 metrics, including time-dependent AUC, Brier score and multiple variations of the C-index.
However, methods were applied with specific hyperparameter sets without tuning, thereby limiting the generalizability of their results.

\paragraph{Comparisons on High-Dimensional Data}
\cite{herrmann2020} performed a large-scale benchmark experiment of survival models on multi-omics high-dimensional data.
Models fall into the following groups: Penalised regression, GBMs, and RSFs.
Comparisons are made with Uno's C and the Integrated Survival Brier Score (ISBS).
The ISBS for all models overlapped with the Kaplan-Meier baseline though all C-indices were significantly higher than the baseline --- however it is not stated how the standard errors for the confidence intervals were derived nor is it stated in the paper if multiple testing correction is applied.
The authors also note that their results should be treated with caution due to the small performance differences and high variability.
%
\cite{spooner2020} also compared machine learning models on high-dimensional data.
In this study GBMs, RSFs, CPH and some extensions thereof were compared.
Models were evaluated by Harrell's C-index only.
The results indicated that all models outperformed CPH when no additional feature selection was used but that there were no significant differences when feature selection was applied to the Cox model.
There were few significant statistical differences between models.
%
\cite{wissel2023systematic} provide a systematic comparison of multi-omics cancer survival models comparing eight DL methods, RSFs and CPH.
They primarily focus on the noise-resistance of these models for high-dimensional settings, finding a general lack thereof when evaluating on Antolini's C-index and ISBS.

\paragraph{Comparisons of Classical Models}

\cite{moghimi-dehkordi2008} compare CPH to AFT models with various distributions.
Out-of-sample measures for comparison are not provided though the AIC produced by the CPH is far higher (and therefore inferior) than those of the parametric models.
Model inspection demonstrated that all models provided similar (non-significantly different) confidence intervals for hazard ratios.
%
\cite{georgousopoulou2015} compared the CPH to a Weibull and Exponential AFT model.
Again no out-of-sample measures were utilized, models were compared by the Cox-Snell residuals and the Bayesian Information Criterion (BIC).
Similarly to Moghimi-dehkordi \etal, hazard ratios produced from all three models were nearly identical.
The authors claim the CPH is inferior to the parametric models though only graphical comparisons are included.
%
\cite{zare2015} provide another comparison of the CPH to AFT models, using Cox-Snell residuals and AIC as their measures of comparison.
Similarly to the previous studies the Cox model has the highest AIC though the plotted Cox-Snell residuals are very similar.
The authors acknowledge no significant differences between the model classes and instead conclude that AFT is a useful and more interpretable alternative.
No significant differences were found between the different AFT parameterizations.
%
\cite{dirick2017} make use of a financial setting to compare the CPH, AFT, flexible Cox models using splines to model the hazard, and mixture cure models.
The authors compare the models using a time-dependent AUC, the mean squared error (MSE), and the mean absolute error (MAE).
Survival times are generated from the CPH with a deterministic composition using quantiles chosen to minimize the MSE and MAE, and it is not clear if this is performed in an unbiased nested resampling manner or after predictions are made.
By averaging the ranking of model performance the authors conclude that CPH with penalized splines outperformed the other models with respect to the chosen metrics.

\cite{habibi2018} performed another experiment on PH and AFT models.
Models were again compared exclusively by the AIC with the PH having the highest result and log-normal AFT the lowest; differences between AFT models were non-significant.
Confidence intervals for hazard ratios were similar (non-significantly different) for all models.

\paragraph{Comparisons of a Novel Model Class}

\cite{luxhoj1997} benchmarked neural networks against CPH in the engineering field of reliability analysis.
The baseline hazard of the Cox model is modeled by splines with a single knot.
The models are compared using the mean squared error on a validation set of sample size 40, of these 40 there are only 9 unique failure times that are used for model testing.
Insufficient information is provided to determine the architecture or training procedure of the neural networks compared.
The MSE difference between the Cox and ANN was 0.003, which is highly unlikely to be a significant difference on a test set of only 40 observations with nine observed events.
%
\cite{ohno-machado1997} also compared CPH to ANNs.
Several Cox models were fit with automated variable selection by backwards elimination.
For each model, survival curves were predicted and for a given patient they were considered dead at a particular time point if the predicted survival curve at the time is less than the \enquote{arbitrary}~
\cite{ohno-machado1997} probability of $0.5$.
The Cox models were compared to a single hidden layer ANN.
This model made probabilistic predictions of death in four time-intervals that were within the predicted time of the Cox models.
The probabilistic predictions from both models were compared with the AUC and its corresponding ROC.
No significant differences in performance were found between the two models.
%
\cite{goli2016a} provide a comprehensive comparison of support vector machine models with CPH as a reference class (Kaplan-Meier is not included).
Models are compared against the C-index and log-rank test, though it is unstated which C-index is utilized.
No model outperformed CPH with respect to the chosen C-index.
%
\cite{jaeger2024} compare a multiple variations of a novel implementation of oblique RSFs (\enquote{aorsf}) to the previous implementation, as well as other RSFs, GBMs, penalized CPH, and ANNs.
They compare methods on 21 datasets, including low- and high-dimensional settings.
Results are analyzed using post-hoc Bayesian ROPE and evaluated using the Index of Prediction Accuracy (IPA)~\citep{kattan2018indexprediction} based on the ISBS and time-dependent C-index~\citep{blanche2013estimatingcomparing}.
They present results relative to the method performing best in their benchmark (\enquote{aorsf-fast} for both measures), with GBMs among the lowest-performing methods.
Only minimal tuning was conducted.

\subsection{Surveys of Survival Models}
\label{sec:lit_survey}

The final class of papers do not perform empirical benchmark experiments but instead survey/review available survival models.
These are therefore only discussed very briefly.
%
\cite{ohno-machado2001} provide an overview to models available for survival analysis from non-parametric estimators and classical models to neural networks.
The review highlights useful applications of the models and their respective limitations.
In particular their Table 1 clearly states advantages and disadvantages of Cox models versus ANNs.
%
\cite{patel2006} compare proportional hazards and accelerated failure time models. This comparison is primarily theoretical and based on model properties, no analytical comparison with measures is provided though comparisons of predicted median survival times are compared to those from a Kaplan-Meier estimator. The authors conclude that AFT models should be considered more often due to simpler interpretation.
%
\cite{wang2019} provide a review of survival analysis models and measures that is strongly recommended here as a precise and comprehensive introduction to the field of machine learning in survival analysis.
The authors provide strong arguments for comparing classical models against one another, though this is not extended to the machine learning setting.
More mathematical detail is provided to the classical setting however a clear and detailed overview is still provided for machine learning models. Some attention is also given to the more complex cases of competing risks and multiple events.
%
\cite{lee2019} provide a short but concise overview to survival analysis models with an emphasis on genetic data and implementation in \Rstats.
Their review covers classical models, penalisation, and many machine learning models.
They provide a clear, practical illustration (but no full benchmark experiment) comparing the models on a real dataset against Harrell's C. No model outperforms CPH.
%
\cite{wiegrebe2024deeplearning} provide a comprehensive overview of deep learning methods for survival analysis and compare methods based on their capabilities regarding various common challenges in survival analysis such as time-varying features, competing events, different censoring types, dimensionality, modality, and interpretability.
Their extensive comparison is also available as a web-based interactive table.

\section{Datasets}\label{app:datasets}

\Cref{tab:datasets} lists datasets included in the benchmark along with their sources and common descriptive statistics.
\Cref{tab:datasets_licenses} lists the licenses declared by the source packages from which the dataset is taken.

\clearpage
\begin{table}[ht!]
    \centering
    \caption{Datasets used in benchmark experiment.}\label{tab:datasets}
    \begin{tabular}{llllllll}
        \toprule
        \textbf{Dataset}$^1$ & \textbf{Cens \%}$^2$ & $n_C^3$ & $n_D^4$ & \textbf{n}$^5$ & \textbf{p}$^6$ & ${n_E}^7$ & \textbf{Package}$^8$ \\ 
        \hline
        aids.id~\citep{dataaidsid} & 60 & 1 & 4 & 467 & 5 & 188 & \pkg{JM}~\citep{pkgjm}  \\
        aids2~\citep{pkgnnet} & 38 & 1 & 3 & 2814 & 4 & 1733 & \pkg{MASS}~\citep{pkgnnet} \\
        bladder0~\citep{databladder0} & 48 & 0 & 3 & 397 & 3 & 206 & \pkg{frailtyHL}~\citep{pkgfrailtyhl} \\
        CarpenterFdaData~\citep{carpenter2002} & 36 & 15 & 11 & 408 & 26 & 262 & \pkg{simPH} \\
        cat\_adoption~\citep{pkgmodeldata}  & 37 & 14 & 4 & 2257 & 18 & 1434 & \pkg{modeldata}~\citep{pkgmodeldata} \\
        channing~\citep{klein2003} & 62 & 1 & 1 & 458 & 2 & 176 & \pkg{KMsurv} \\
        check\_times~\citep{pkgmodeldata}  & 1 & 22 & 0 & 13626 & 22 & 13523 & \pkg{modeldata} \\
        child~\citep{pkgeha} & 79 & 1 & 3 & 26574 & 4 & 5616 & \pkg{eha}~\citep{pkgeha} \\
        colrec~\citep{pkgrelsurv} & 17 &  3 & 2 & 5578 & 13 & 4602 & \pkg{relsurv}~\citep{pkgrelsurv} \\
        cost~\citep{datacost} & 22 & 3 & 10 & 518 & 13 & 404 & \pkg{pec}~\citep{pkgpec} \\
        dataFTR~\citep{dataftrstr} & 86 & 0 & 2 & 2206 & 2 & 300 & \pkg{RISCA}~\citep{pkgrisca} \\
        dataSTR~\citep{dataftrstr} & 82 & 0 & 4 & 546 & 4 & 101 & \pkg{RISCA} \\
        e1684~\citep{datae1684} & 31 & 1 & 2 & 284 & 3 & 196 & \pkg{smcure}~\citep{pkgsmcure} \\
        flchain~\citep{dataflchain} & 72 & 4 & 3 & 7871 & 7 & 1082 & \pkg{survival} \\
        gbsg~\citep{katzman2018} & 43 & 3 & 4 & 2232 & 7 & 1267 & \pkg{pycox}~\citep{pkgpycox} \\
        grace~\citep{dataapplied} & 68 & 4 & 2 & 1000 & 6 & 324 & \pkg{mlr3proba}~\citep{pkgmlr3proba} \\
        hdfail~\citep{pkgfrailtysurv} & 94 & 1 & 4 & 52422 & 5 & 2885 & \pkg{frailtySurv}~\citep{pkgfrailtysurv} \\
        kidtran~\citep{klein2003} & 84 & 1 & 3 & 863 & 4 & 140 & \pkg{KMsurv} \\
        liver~\citep{dataliver} & 40 & 1 & 1 & 488 & 2 & 292 & \pkg{joineR}~\citep{pkgjoiner} \\
        lung~\citep{datalung} & 28 & 5 & 3 & 167 & 8 & 120 & \pkg{survival} \\ 
        metabric~\citep{katzman2018} & 42 & 5 & 4 & 1903 & 9 & 1103 & \pkg{pycox} \\ 
        mgus~\citep{datamgus} & 6 & 6 & 1 & 176 & 7 & 165 & \pkg{survival} \\ 
        nafld1~\citep{datanafld1} & 92 & 4 & 1 & 12446 & 5 & 1018 & \pkg{survival} \\ 
        nwtco~\citep{datanwtco} & 86 & 1 & 2 & 4028 & 3 & 571 & \pkg{survival} \\
        ova~\citep{dataova} & 26 & 1 & 4 & 358 & 5 & 266 & \pkg{dynpred} \\
        rdata~\citep{pkgrelsurv} & 47 & 1 & 3 & 1040 & 4 & 547 & \pkg{relsurv}~\citep{pkgrelsurv} \\
        std~\citep{klein2003} & 60 & 3 & 18 & 877 & 21 & 347 & \pkg{KMsurv} \\
        support~\citep{katzman2018} & 32 & 10 & 4 & 8873 & 14 & 2705 & \pkg{pycox} \\ 
        tumor~\citep{pkgpammtools} & 52 & 1 & 6 & 776 & 7 & 375 & \pkg{pammtools} \\
        uis~\citep{datauis} & 19 & 7 & 5 & 575 & 12 & 464 & \pkg{quantreg}~\citep{pkgquantreg} \\
        veteran~\citep{kalbfleisch2011} & 7 & 3 & 3 & 137 & 6 & 128 & \pkg{survival} \\
        wa\_churn~\citep{pkgmodeldata}  & 73 & 8 & 10 & 7032 & 18 & 1869 & \pkg{modeldata} \\
        wbc1~\citep{datawbc} & 43 & 2 & 0 & 190 & 4 & 109 & \pkg{dynpred} \\
        whas~\citep{dataapplied} & 48 & 3 & 6 & 481 & 9 & 249 & \pkg{mlr3proba} \\
        \bottomrule
    \end{tabular}
    \label{tab:real}
    \begin{tablenotes}
        \footnotesize
        \item 1. Dataset ID and citation.
        \item 2. Proportion of censoring in the (modified) dataset, rounded to nearest percentage point.
        \item 3-4. Number of continuous and discrete features respectively before recoding.
        \item 5-6. Total number of observations and features respectively after alterations described above but before sub-sampling.
        \item 7. Number of observed events in dataset.
        \item 8. Package in which the dataset is included.
    \end{tablenotes}
\end{table}

\clearpage
\begin{table}[ht!]
\label{tab:datasets_licenses}
\caption{Software licenses of the R and Python packages used as sources for datasets in this benchmark.}
\centering
\begin{tabular}[t]{ll}
\toprule
License & Packages\\
\midrule
BSD-2 & pycox\\
GPL & relsurv\\
GPL ($\geq$ 2) & jm, eha, pec, RISCA, dynpred, quantreg\\
GPL ($\geq$ 3) & KMsurv\\
GPL-2 & smcure\\
GPL-2 $\|$ GPL-3 & nnet\\
GPL-3 & simPH, joineR\\
LGPL ($\geq$ 2) & survival\\
LGPL-2 & frailtySurv\\
LGPL-3 & mlr3proba\\
MIT & modeldata, pammtools\\
Unlimited & frailtyHL\\
\bottomrule
\end{tabular}
\end{table}

\section{Models and Configurations}\label{app:models}  

\Cref{tab:models} lists all the compared algorithms, along with their R package name, and prediction types.
The horizontal lines separate the models into different groups: (1) Baseline learners (2) Classical parametric models including penalized versions (3) Tree-based methods, including random survival forests and Trees (4) Boosting algorithms including gradient- and likelihood-boosting (5) Support Vector Machines (SVMs).
\Cref{tab:config} shows hyperparameter search spaces and non-default parameter values as well as common pre-processing requirements.

\clearpage
\thispagestyle{empty}
\onecolumn
\begin{landscape}
    \begin{longtable}{ccccccc}
        \caption{Models used for benchmarking with associated packages and prediction types.}\label{tab:models} \\
        \toprule
        \multicolumn{3}{c}{\textbf{Model Information}}  & \multicolumn{3}{c}{\textbf{Prediction Types}}  \\
        \cmidrule(lr){1-3} \cmidrule(lr){4-7}
        \textbf{Model Name}$^1$   & \textbf{Learner}$^2$  & \textbf{Package}$^3$   & \textbf{distr}$^4$ & \textbf{crank}$^5$ & \textbf{lp}$^6$ \\
        \midrule
        \endfirsthead
        \caption[]{(continued)}  \\
        \toprule
        \multicolumn{3}{c}{\textbf{Model Info}}   & \multicolumn{4}{c}{\textbf{Prediction Types}} \\
        \cmidrule(lr){1-3} \cmidrule(lr){4-7}
        \textbf{Model Name}$^1$   & \textbf{Learner}$^2$  & \textbf{Package}$^3$   & \textbf{distr}$^4$ & \textbf{crank}$^5$ & \textbf{lp} ($\eta$)$^6$ \\
        \midrule
        \endhead
        
        \hline
        \multicolumn{7}{r}{\emph{Continued on next page...}} \\
        \midrule
        \endfoot
        \bottomrule
        \endlastfoot
        
        Kaplan-Meier (KM)~\citep{kaplanmeier1958}                   & kaplan       & survival         & \ding{51}           & ExpMort            & \ding{53}       \\
        Nelson-Aalen (NEL)~\citep{aalen1978}                        & nelson       & survival         & \ding{51}           & ExpMort            & \ding{53}       \\
        Akritas Estimator (AK)~\citep{akritas1994}                  & akritas      & survivalmodels   & \ding{51}           & ExpMort            & \ding{53}       \\
        \hline 
        Cox PH (CPH)~\citep{cox1972}                                & coxph        & survival         & \ding{51} (Breslow) & lp                 & \ding{51}       \\
        CV Regularized CPH (GLMN)~\citep{simon2011}                 & cv\_glmnet   & glmnet           & \ding{51} (Breslow) & lp                 & \ding{51}       \\
        Penalized (Pen)~\citep{goeman2010l1penalized}               & penalized    & penalized        & \ding{51} (Breslow) & ExpMort            & \ding{53}       \\
        Parametric (AFT) ~\citep{kalbfleisch2011}                   & parametric   & survival         & AFT                 & lp                 & \ding{51}       \\
        Flexible Splines (Flex)~\citep{roystonparmar2002}           & flexible     & flexsurv         & \ding{51}           & lp                 & \ding{51}       \\
        \hline
        Random Survival Forest (RFSRC)~\citep{ishwaran2008}         & rfsrc         & randomForestSRC & \ding{51}           & ExpMort            & \ding{53}       \\
        Random Survival Forest (RAN)~\citep{ishwaran2008,pkgranger} & ranger        & ranger          & \ding{51}           & ExpMort            & \ding{53}       \\
        Conditional Inference Forest (CIF)~\citep{hothorn2006}      & cforest       & partykit        & \ding{51}           & ExpMort            & \ding{53}       \\
        Oblique Random Survival Forest (ORSF)~\citep{jaeger2024}    & orsf          & aorsf           & \ding{51}           & ExpMort            & \ding{53}       \\
        Relative Risk Tree (RRT)~\citep{breiman1984}                & rpart         & rpart           & \ding{53}           & \ding{51}          & \ding{53}       \\
        \hline
        Model-Based Boosting (MBSTCox)~\citep{buhlmann2003boostingl2}  & mboost        & mboost       & \ding{51} (Breslow) & lp                 & \ding{51}       \\
        Model-Based Boosting (MBSTAFT)~\citep{buhlmann2003boostingl2}  & mboost        & mboost       & \ding{53}           & lp                 & \ding{51}       \\

        CoxBoost (CoxB)~\citep{binder2008allowing}                  & cv\_coxboost  & CoxBoost        & \ding{51} (Breslow) & lp                 & \ding{51}       \\
        XGBoost (XGBCox)~\citep{chen2016xgboost}                    & xgboost       & xgboost         & Breslow             & lp                 & \ding{51}       \\
        XGBoost (XGBAFT)~\citep{barnwal2022}                        & xgboost       & xgboost         & \ding{53}           & lp                 & \ding{51}       \\
        \hline
        SSVM-Hybrid (SSVM)~\citep{vanbelle2011supportvector}        & svm           & survivalsvm     &                     & \ding{51}          & \ding{53}       \\
    \end{longtable}
    \begin{tablenotes}
        \footnotesize
        \item 1. Identifier for the algorithm. Model abbreviations in parentheses are used in results.
        \item 2. Learner ID in \pkg{mlr3}.
        \item 3. Package in which the learner is implemented. Most learners are available in \pkg{mlr3extralearners}, with KM, RRT, and CPH provided via \pkg{mlr3proba}, where existing R implementations are wrapped into the \pkg{mlr3} framework. AK is natively implemented in \pkg{survivalmodels}.
        \item 4. \code{distr} predict type in \pkg{mlr3proba} is the probabilistic prediction.
        A check (\ding{51}) incidcates that the distribution is provided directly by the package, somtimes via the Breslow estimator (in parentheses).
        Parametric (AFT) models construct the distribution internally in \pkg{survivalmodels}, while XGBCox relies on \pkg{mlr3proba}’s \code{breslow()} function since \pkg{xgboost} does not estimate distributions by default.
        A cross (\ding{53}) means the prediction is not available.
        \item 5. \code{crank} predict type in \pkg{mlr3proba} is the continuous ranking prediction. 
        A check (\ding{51}) indicates that the ranking is being predicted directly by the package. `ExpMort' stands for expected mortality and is a risk score composed from the predicted survival distribution (\code{distr}). `lp' represents the ranking being identical to the predicted linear predictor. 
        \item 6. \code{lp} predict type in \pkg{mlr3proba} is the linear predictor prediction. A check (\ding{51}) represents the linear predictor being predicted directly by the package whereas a cross (\ding{53}) means the prediction is not available (and cannot be composed).
    \end{tablenotes}
\end{landscape}

\begin{landscape}
    \begin{longtable}{ccccc}
        \caption{Hyper-parameter search-spaces for tuning and non-default configurations for models.}\label{tab:config}\\
        \toprule
        \textbf{Model} & \textbf{Hyper-parameters}$^1$ & \textbf{Values}$^2$ & \textbf{Standardize}$^3$ & \textbf{Encode}$^4$ \\
        \midrule
        \endfirsthead
        \caption[]{(continued)}\\
        \toprule
        \textbf{Model} & \textbf{Hyper-parameters}$^1$ & \textbf{Values}$^2$ & \textbf{Standardize}$^3$ & \textbf{Encode}$^4$ \\
        \midrule
        \endhead
        
        \hline
        \multicolumn{5}{r}{\emph{Continued on next page...}} \\ \midrule
        \endfoot
        \bottomrule
        \endlastfoot
        
        KM & - & - & \ding{53} & \ding{53} \\ \hline
        NEL & - & - & \ding{53} & \ding{53} \\ \hline
        AK & lambda & $[0, 1]$ & \ding{53} & \ding{53} \\ \hline
        CPH & - & - & \ding{53} & \ding{53} \\ \hline
        GLMN & alpha & $[0, 1]$ & \ding{53} & \ding{51} \\
        \hline
        Pen & \makecell{lambda1 \\ lambda2} & \makecell{$2^{[-10, 10]}$ \\ $2^{[-10, 10]}$} & \ding{53} & \ding{53} \\
        \hline
        AFT & dist & \{weibull, lognormal, loglogistic\} & \ding{53} & \ding{51} \\
        \hline
        Flex & \makecell{k} & \makecell{$\{1,...,10\}$} & \ding{53} & \ding{53}  \\
        \hline
        RFSRC & \makecell{splitrule \\ ntree \\ mtry \\ nodesize \\ samptype \\ sampsize} & \makecell{\{logrank, bs.gradient\} \\ 1000 \\ $\{1,...,p\}$ \\ $\{1,...,50\}$ \\ \{swr, swor\} \\ $[0,1]$} & \ding{53} & \ding{53} \\
        \hline
        RAN & \makecell{splitrule \\ num.trees \\ mtry \\ min.node.size \\ replace \\ fraction} & \makecell{\{logrank,C,maxstat\} \\ 1000 \\ $\{1,...,p\}$ \\ $\{1,...,50\}$ \\ \{TRUE, FALSE\} \\ $[0,1]$} & \ding{53} & \ding{53} \\
        \hline
        CIF & \makecell{ntree \\ mtry \\ minsplit \\ mincriterion \\ replace \\ fraction} & \makecell{1000 \\ $\{1,...,p\}$ \\ \{1,...,50\} \\ $[0,1]$ \\ \{TRUE, FALSE\} \\ $[0,1]$} & \ding{53} & \ding{53} \\
        \hline
        ORSF & \makecell{control\_type \\ n\_tree \\ mtry \\ leaf\_min\_events \\ min\_obs\_to\_split\_node \\ alpha} & \makecell{fast \\ 1000 \\ $\{1,...,p\}$ \\ \{5,...,50\} \\ min\_events\_to\_split\_node + 5 \\ $(0,1)$} & \ding{53} & \ding{53} \\
        \hline
        RRT & \makecell{minbucket} & \makecell{$\{5,...,50\}$} & \ding{53} & \ding{53} \\
        \hline
        MBSTCox & \makecell{family \\ mstop \\ nu \\ baselearner} & \makecell{coxph \\ $\{10,...,5000\}$ \\ $(0, 0.1]$ \\ \{bols, btree\}} & \ding{53} & \ding{53} \\
        MBSTAFT & \makecell{family \\ mstop \\ nu \\ baselearner} & \makecell{\{gehan, weibull\} \\ $\{10,...,5000\}$ \\ $(0, 0.1]$ \\ \{bols, btree\}} & \ding{53} & \ding{53} \\
        \hline
        CoxB & \makecell{penalty \\ maxstepno \\ K} & \makecell{optimCoxBoostPenalty \\ 5000 \\ 3} & \ding{53} & \ding{51} \\
        \hline
        XGBCox & \makecell{objective \\ tree\_method \\ booster \\ max\_depth \\ subsample \\ colsample\_bytree \\ early\_stopping\_rounds \\ eta \\ grow\_policy} & \makecell{survival:cox \\ hist \\ gbtree \\ \{1,...,20\} \\ $[0,1]$ \\ $(0,1]$ \\ 50 \\ $[10^{-5}, 10^5]$ \\ \{depthwise, lossguide\}} & \ding{53} & \ding{51} \\
        \hline
        XGBAFT & \makecell{objective \\ tree\_method \\ booster \\ max\_depth \\ subsample \\ colsample\_bytree \\ early\_stopping\_rounds \\ eta \\ grow\_policy \\ aft\_loss\_distribution \\ aft\_loss\_distribution\_scale} & \makecell{survival:aft\\ hist \\ gbtree \\ \{1,...,20\} \\ $[0,1]$ \\ $(0,1]$ \\ 50 \\ $[10^{-5}, 10^5]$ \\ \{depthwise, lossguide\} \\  \{normal, logistic, extreme\} \\ $[0.5, 2]$ } & \ding{53} & \ding{51} \\
        \hline
        SSVM$^5$ & \makecell{type \\ diff.meth \\ gamma.mu \\ kernel \\ kernel.pars} & \makecell{hybrid \\ makediff3 \\ $([2^{-10}, 2^{10}], [2^{-10}, 2^{10}])$ \\ \{lin\_kernel, rbf\_kernel, add\_kernel\} \\ $[2^{-5},2^5]$} & \ding{51} & \ding{51} \\
    \end{longtable}
    \begin{tablenotes}
        \footnotesize
        \item $^1$ Hyper-parameters for model tuning. The choice of hyper-parameters are largely informed by recommendations from the model author and subsequent papers exploring optimization. A `-' indicates no tuning is performed.
        \item $^2$ Value ranges for the respective hyper-parameters to tune over. Omitted parameters use the package defaults.
        \item $^3$ Pre-processing of covariates by scaling to unit variance and centering to zero mean.
        A check (\ding{51}) indicates this step is performed before training the model, and a cross (\ding{53}) if not.
        \item $^4$ Pre-processing of covariates by treatment encoding with \code{model.matrix}.
        A check (\ding{51}) indicates this step is performed before training the model, and a cross (\ding{53}) if not.
    \end{tablenotes}
\end{landscape}

\section{Implementation, Reproducibility, Accessibility}\label{app:reprodubility}

\textbf{Platform}
All experiments were conducted on \Rstats 4.4.3 (2025-02-28) -- \enquote{Trophy Case} on the Beartooth Computing Environment \citep{beartooth}.

\textbf{Reproducibility and Accessibility}
Seeds were set with L'Ecuyer's random number generator~\citep{lecuyer1999} to ensure reproducible results.
All code required to run the experiments, as well as the results, are freely available in a public GitHub repository (\url{https://github.com/slds-lmu/paper_2023_survival_benchmark}.
Software, packages, and version numbers that were utilized to conduct benchmark and analysis are listed below.
We also employ the \pkg{renv} R package~\citep{pkgrenv} to record and restore package dependencies to further aid reproducibility.

\textbf{Packages}
All code is implemented in \Rstats and the experiment was run with \pkg{batchtools}~\citep{lang2017batchtoolstools}.
Learners are implemented in \pkg{mlr3proba}~\citep{pkgmlr3proba}, \pkg{mlr3extralearners}~\citep{pkgmlr3extralearners} and \pkg{survivalmodels}~\citep{pkgsurvivalmodels}.
Tuning is implemented in \pkg{mlr3tuning}~\citep{pkgmlr3tuning}.
Benchmarking functionality is implemented in \pkg{mlr3}~\citep{pkgmlr3}.
Statistical benchmark analysis is implemented in \pkg{mlr3benchmark}~\citep{pkgmlr3benchmark}.
Compositions and pre-processing steps are implemented through \pkg{mlr3pipelines}~\citep{pkgmlr3pipelines}.
The implementing packages for all algorithms are given in \Cref{tab:models} and all measures are implemented in \pkg{mlr3proba}.

\section{Results}\label{app:results}

The following figures show boxplots of the respective evaluation measure across all outer resampling iterations.
Results are also available following links on the GitHub repository at \url{https://github.com/slds-lmu/paper_2023_survival_benchmark}.

As described in \Cref{sec:results}, we provide aggregated results based on average ranks across all datasets as boxplots with two versions for discrimination measures and three different scaling options for scoring rules:

\begin{enumerate}
    \item Raw scores as produced by the evaluation measures, e.g. \enquote{ISBS},
    \item Explained Residual Variation (ERV) scores, e.g. \enquote{ISBS [ERV]},
    \item Scaled scores, e.g. \enquote{ISBS [Scaled]}, which scales raw scores such that 0 is the score achieved by KM and 1 is the score achieved by the best-performing model within the given combination of dataset and tuning- and evaluation measure.
\end{enumerate}

Since the discrimination measures Harrell's C and Uno's C are already scaled from 0 (worst) to 1 (best) with KM achieving a score of 0.5 by design, the ERV option is omitted and only options 1) and 3) are presented.

\subsection{Discrimination Measures}

\subsubsection{Raw Scores}
Boxplots of raw evaluation scores using discrimination measures for tuning (Harrell's C)
\begin{figure}[ht]
    \centering
    \includegraphics[width=0.7\textwidth]{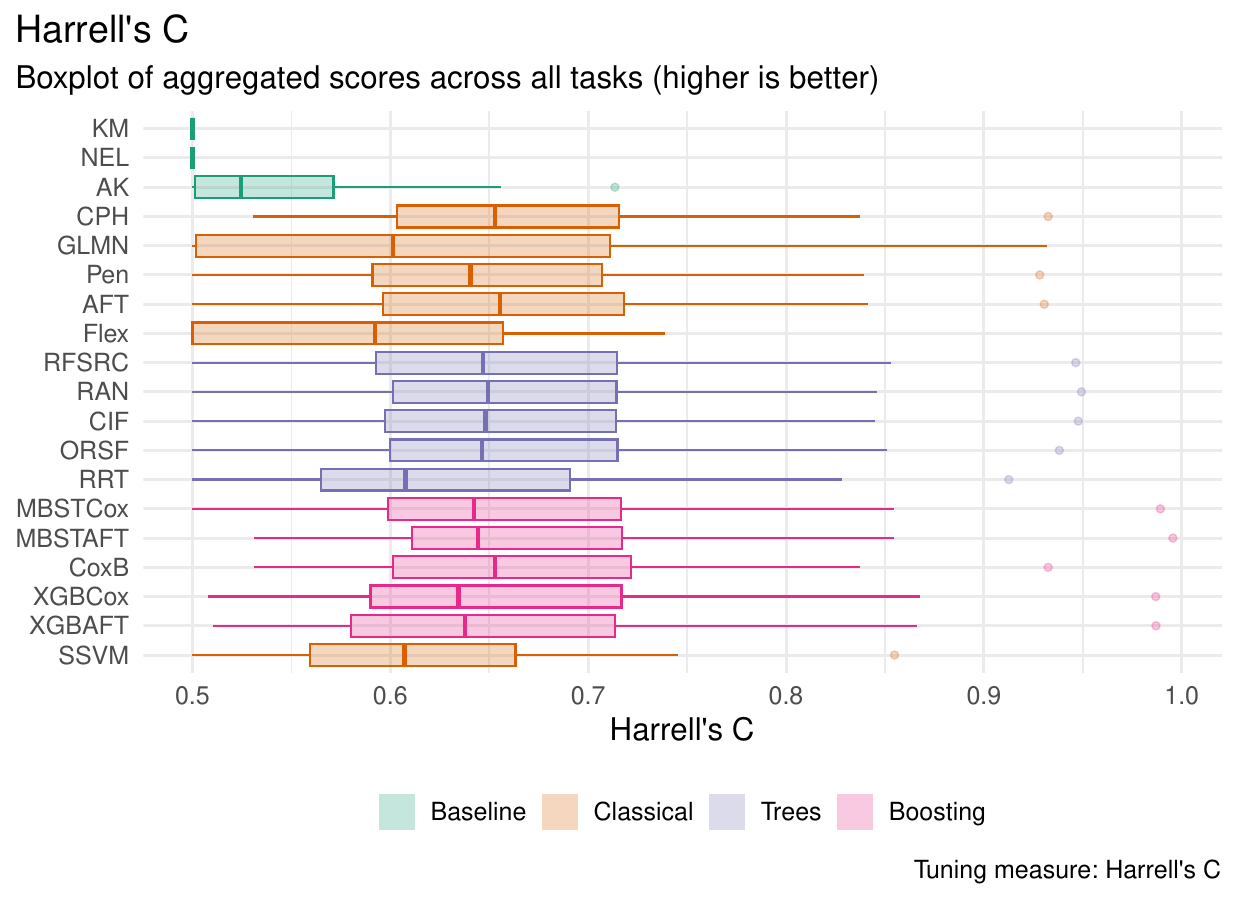}
    \caption{Learners tuned and evaluated with Harrell's C}
\end{figure}

\begin{figure}[ht]
    \centering
    \includegraphics[width=0.7\textwidth]{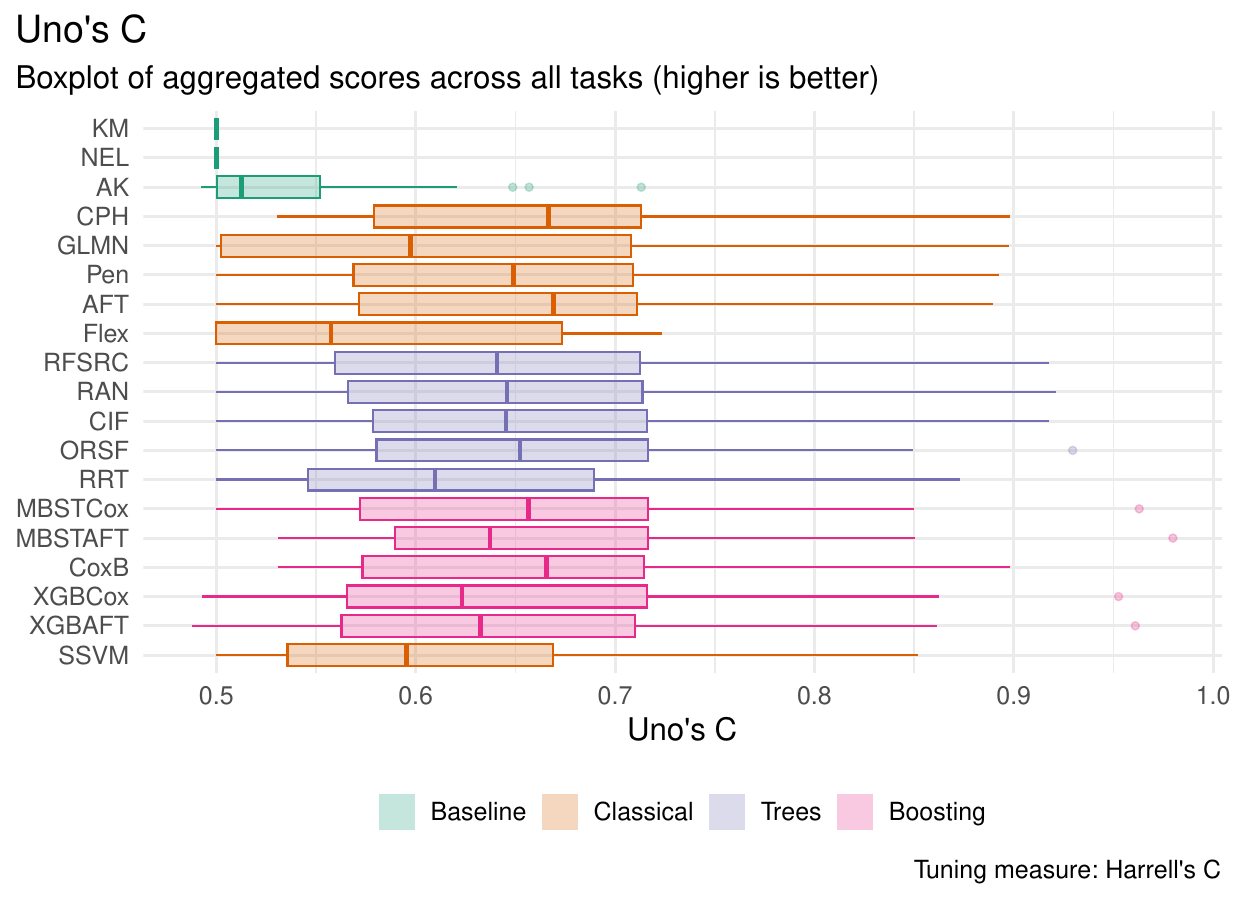}
    \caption{Learners tuned for Harrell's C and evaluated with Uno's C}
\end{figure}

\clearpage
\subsubsection{Scaled}

\begin{figure}[ht]
    \centering
    \includegraphics[width=0.7\textwidth]{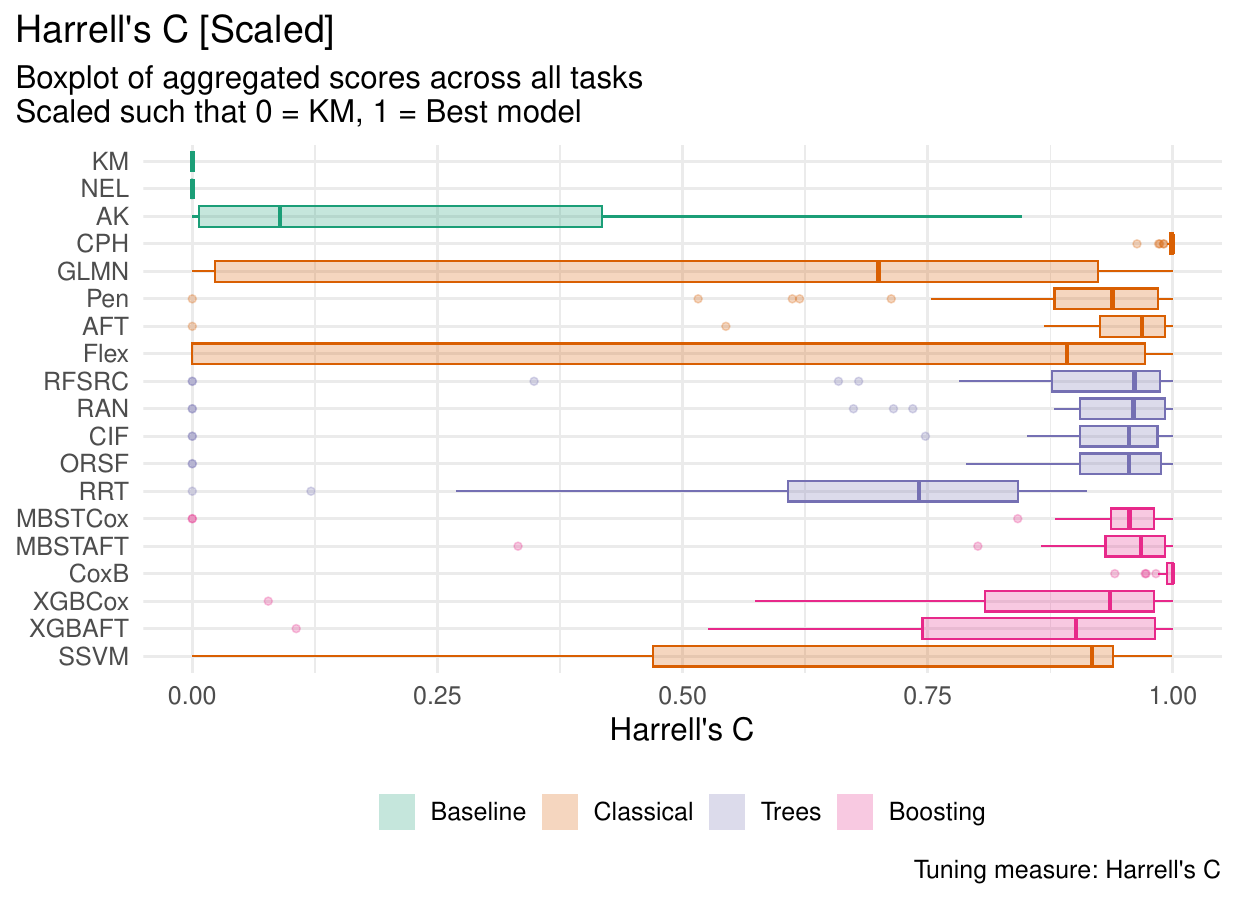}
    \caption{Learners tuned and evaluated with Harrell's C (scaled scores)}
\end{figure}

\begin{figure}[ht]
    \centering
    \includegraphics[width=0.7\textwidth]{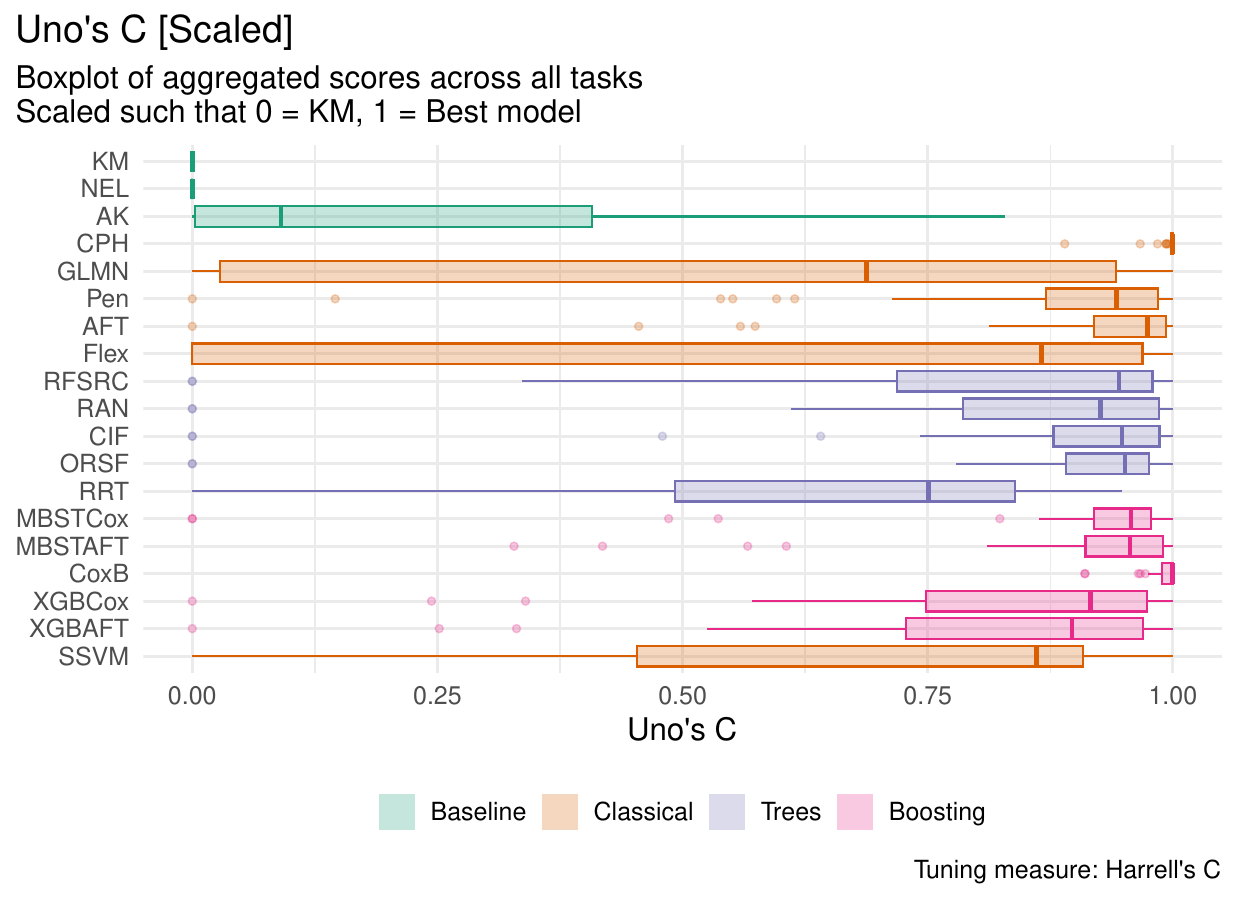}
    \caption{Learners tuned for Harrell's C and evaluated with Uno's C (scaled scores)}
\end{figure}

\clearpage
\subsection{Scoring Rules}

\subsubsection{Raw Scores}

\begin{figure}[ht]
    \centering
    \includegraphics[width=0.7\textwidth]{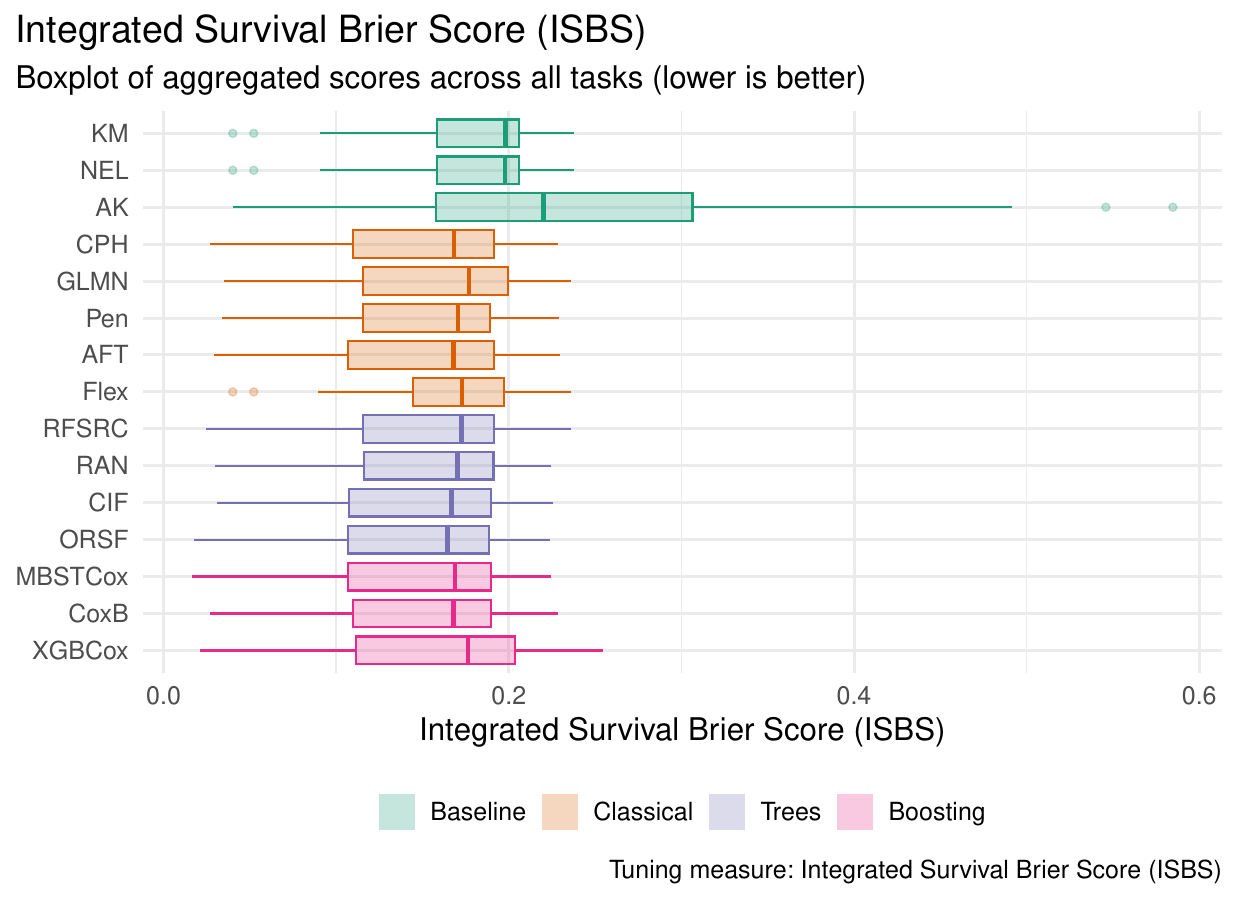}
    \caption{Learners tuned for ISBS and evaluated with ISBS (raw scores)}
\end{figure}

\begin{figure}[ht]
    \centering
    \includegraphics[width=0.7\textwidth]{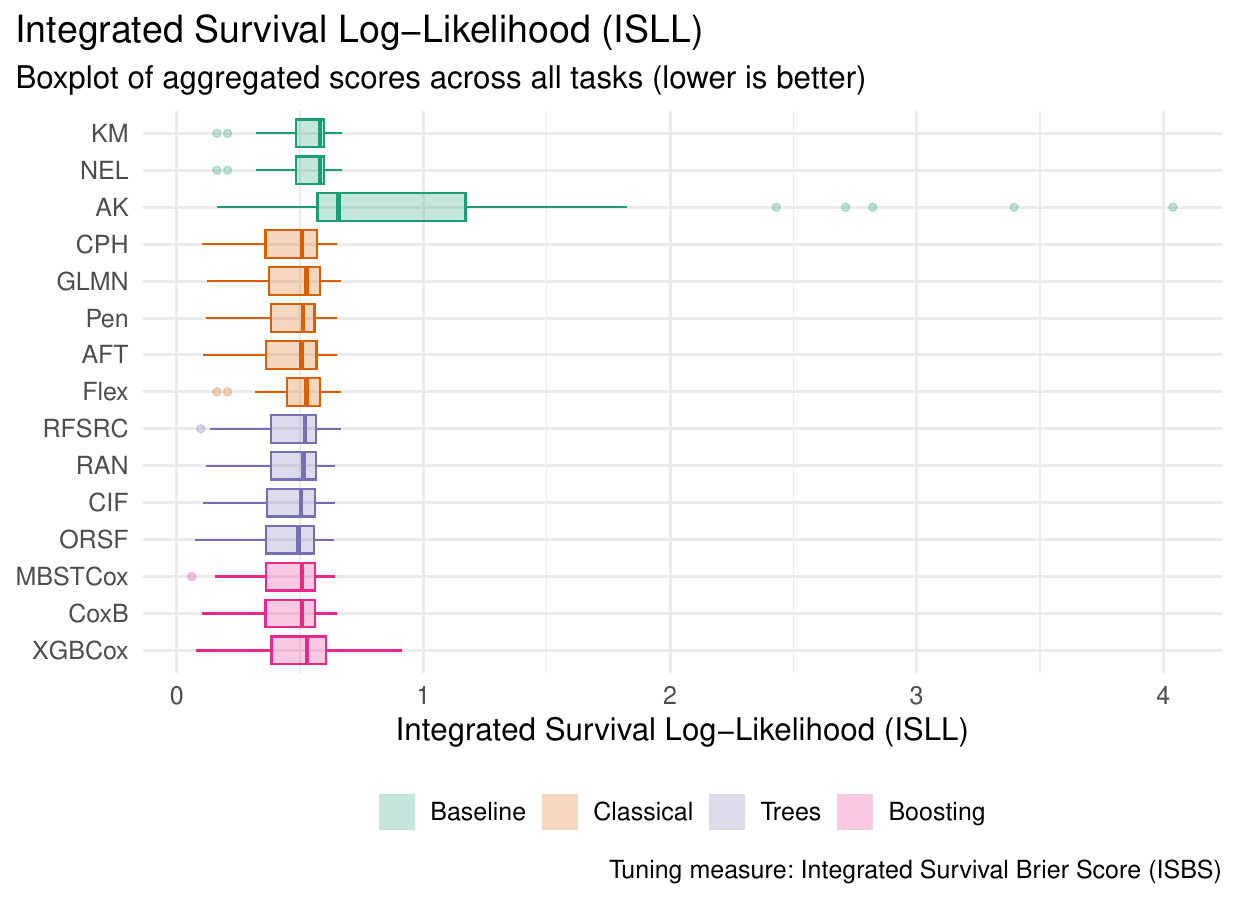}
    \caption{Learners tuned for ISBS and evaluated with ISLL (raw scores)}
\end{figure}

\clearpage
\subsubsection{ERV}

\begin{figure}[ht]
    \centering
    \includegraphics[width=0.7\textwidth]{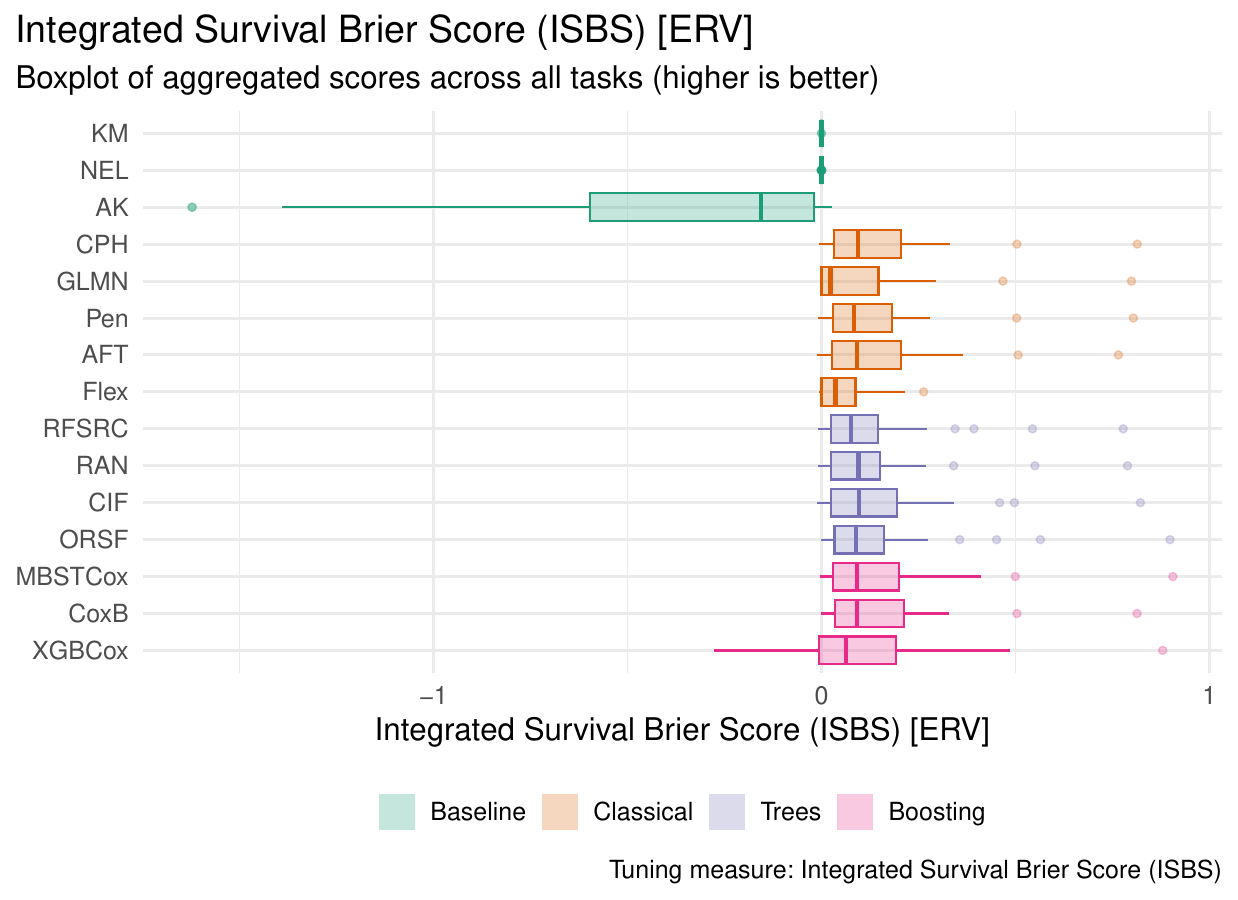}
    \caption{Learners tuned for ISBS and evaluated with ISBS (ERV)}
\end{figure}

\begin{figure}[ht]
    \centering
    \includegraphics[width=0.7\textwidth]{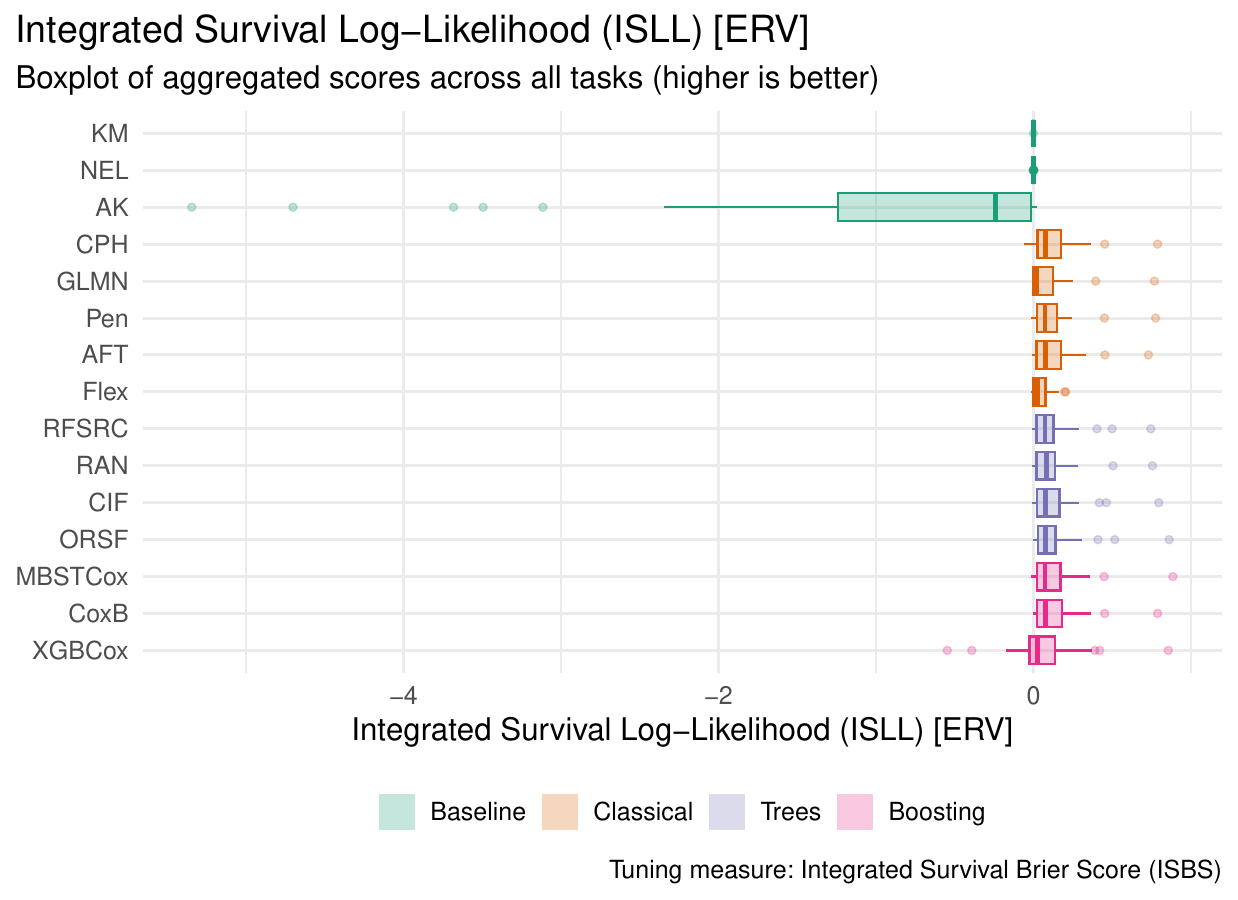}
    \caption{Learners tuned for ISBS and evaluated with ISLL (ERV)}
\end{figure}

\clearpage
\subsubsection{Scaled}

\begin{figure}[ht]
    \centering
    \includegraphics[width=0.7\textwidth]{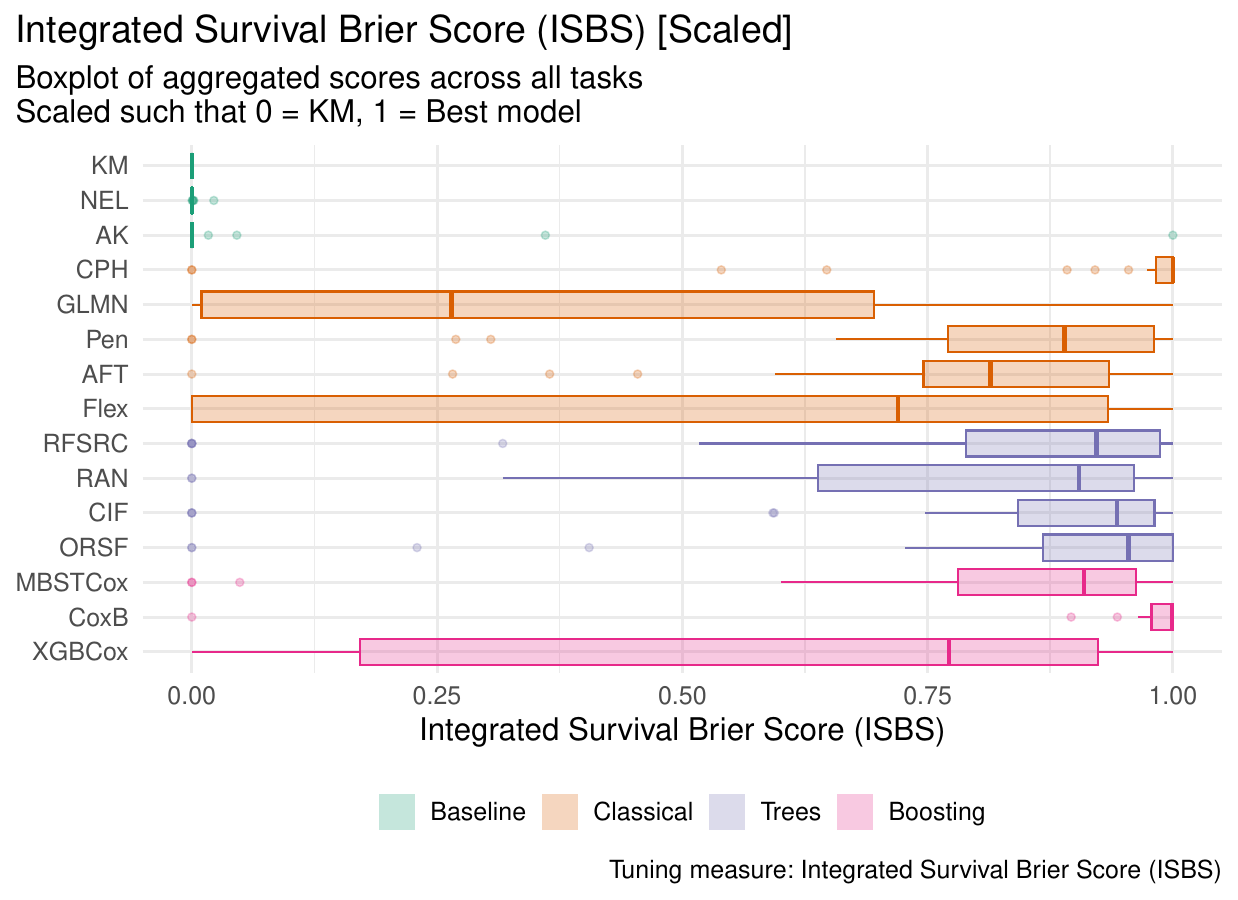}
    \caption{Learners tuned for ISBS and evaluated with ISBS (scaleds)}
\end{figure}

\begin{figure}[ht]
    \centering
    \includegraphics[width=0.7\textwidth]{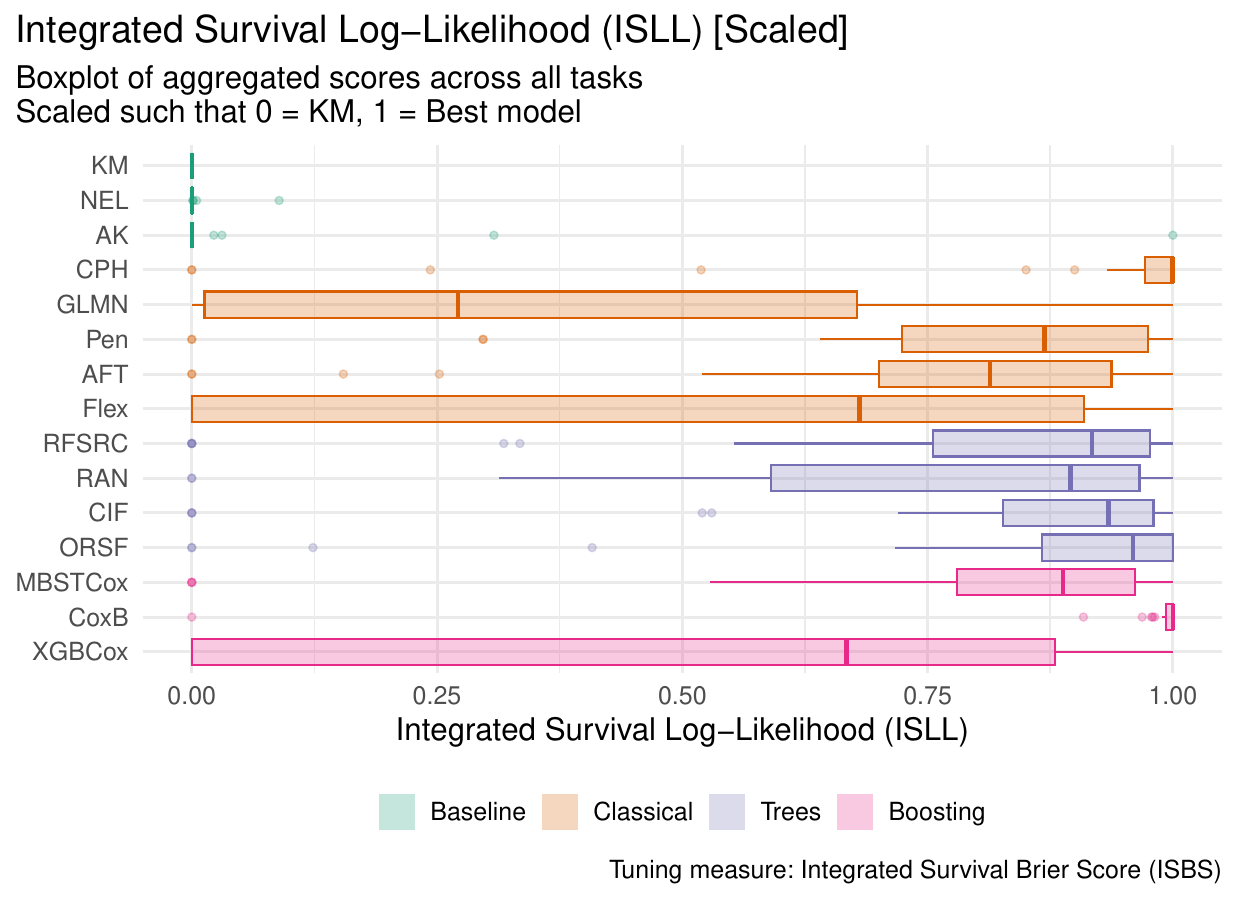}
    \caption{Learners tuned for ISBS and evaluated with ISLL (scaled)}
\end{figure}

\subsection{Errors}\label{app:errors}

During the many computational steps performed during this benchmark, software errors are inevitably bound to happen.
As we noted in \Cref{sec:exp}, we impute missing evaluation scores in resampling folds using the score of the Kaplan-Meier estimator, which affects the results presented in this paper.

Table \ref{tab:errors} counts the number of errors encountered for each model, dataset, and tuning measures uses in the benchmark per outer resampling fold (up to five).
We note that particularly the tasks hdfail and child have caused the majority of the runtime- or memory-related errors here due to their large sample sizes and number of unique time points.

\begin{longtable}[t]{llllr}
\caption{Number of errors per outer resampling iteration (up to 30), separated by model, dataset, and tuning measure.\label{tab:errors}}\\
\toprule
Model & Dataset & Harrell's C & ISBS & Total Errors\\
\midrule
AK & CarpenterFdaData & 0 (0\%) & 1 (3.3\%) & 1\\
AK & channing & 1 (3.3\%) & 1 (3.3\%) & 2\\
AK & child & 3 (100\%) & 3 (100\%) & 6\\
AK & e1684 & 3 (10\%) & 0 (0\%) & 3\\
AK & hdfail & 3 (100\%) & 3 (100\%) & 6\\
AK & lung & 8 (26.7\%) & 0 (0\%) & 8\\
AK & uis & 2 (6.7\%) & 0 (0\%) & 2\\
AK & veteran & 3 (10\%) & 0 (0\%) & 3\\
GLMN & bladder0 & 1 (3.3\%) & 0 (0\%) & 1\\
GLMN & channing & 0 (0\%) & 1 (3.3\%) & 1\\
GLMN & check\_times & 2 (66.7\%) & 0 (0\%) & 2\\
GLMN & cost & 12 (40\%) & 0 (0\%) & 12\\
GLMN & dataSTR & 2 (6.7\%) & 0 (0\%) & 2\\
GLMN & hdfail & 0 (0\%) & 3 (100\%) & 3\\
GLMN & std & 6 (20\%) & 0 (0\%) & 6\\
GLMN & uis & 4 (13.3\%) & 0 (0\%) & 4\\
GLMN & veteran & 0 (0\%) & 14 (46.7\%) & 14\\
GLMN & wbc1 & 0 (0\%) & 4 (13.3\%) & 4\\
Pen & aids.id & 1 (3.3\%) & 9 (30\%) & 10\\
Pen & bladder0 & 8 (26.7\%) & 0 (0\%) & 8\\
Pen & channing & 1 (3.3\%) & 0 (0\%) & 1\\
Pen & check\_times & 3 (100\%) & 3 (100\%) & 6\\
Pen & cost & 3 (10\%) & 0 (0\%) & 3\\
Pen & dataSTR & 11 (36.7\%) & 3 (10\%) & 14\\
Pen & hdfail & 0 (0\%) & 2 (66.7\%) & 2\\
Flex & aids.id & 0 (0\%) & 10 (33.3\%) & 10\\
Flex & check\_times & 3 (100\%) & 3 (100\%) & 6\\
Flex & child & 3 (100\%) & 3 (100\%) & 6\\
Flex & dataFTR & 2 (6.7\%) & 0 (0\%) & 2\\
Flex & hdfail & 3 (100\%) & 3 (100\%) & 6\\
Flex & lung & 9 (30\%) & 0 (0\%) & 9\\
Flex & nafld1 & 14 (93.3\%) & 14 (93.3\%) & 28\\
Flex & nwtco & 15 (100\%) & 15 (100\%) & 30\\
Flex & support & 3 (100\%) & 3 (100\%) & 6\\
Flex & wa\_churn & 15 (100\%) & 15 (100\%) & 30\\
RFSRC & check\_times & 3 (100\%) & 3 (100\%) & 6\\
RFSRC & child & 3 (100\%) & 3 (100\%) & 6\\
RFSRC & colrec & 1 (33.3\%) & 2 (66.7\%) & 3\\
RFSRC & nafld1 & 2 (13.3\%) & 1 (6.7\%) & 3\\
RFSRC & support & 2 (66.7\%) & 3 (100\%) & 5\\
RAN & check\_times & 3 (100\%) & 2 (66.7\%) & 5\\
RAN & child & 3 (100\%) & 3 (100\%) & 6\\
RAN & cost & 1 (3.3\%) & 0 (0\%) & 1\\
RAN & hdfail & 1 (33.3\%) & 1 (33.3\%) & 2\\
RAN & mgus & 0 (0\%) & 2 (6.7\%) & 2\\
RAN & nafld1 & 4 (26.7\%) & 9 (60\%) & 13\\
CIF & child & 3 (100\%) & 3 (100\%) & 6\\
CIF & hdfail & 3 (100\%) & 3 (100\%) & 6\\
ORSF & child & 3 (100\%) & 3 (100\%) & 6\\
ORSF & cost & 1 (3.3\%) & 0 (0\%) & 1\\
ORSF & gbsg & 1 (6.7\%) & 0 (0\%) & 1\\
ORSF & hdfail & 3 (100\%) & 3 (100\%) & 6\\
ORSF & nafld1 & 1 (6.7\%) & 9 (60\%) & 10\\
ORSF & uis & 0 (0\%) & 1 (3.3\%) & 1\\
ORSF & veteran & 1 (3.3\%) & 0 (0\%) & 1\\
RRT & dataFTR & 5 (16.7\%) & 0 (0\%) & 5\\
RRT & lung & 8 (26.7\%) & 0 (0\%) & 8\\
RRT & metabric & 7 (46.7\%) & 0 (0\%) & 7\\
RRT & nwtco & 7 (46.7\%) & 0 (0\%) & 7\\
RRT & ova & 3 (10\%) & 0 (0\%) & 3\\
RRT & tumor & 3 (10\%) & 0 (0\%) & 3\\
MBSTCox & child & 3 (100\%) & 3 (100\%) & 6\\
MBSTCox & dataSTR & 0 (0\%) & 1 (3.3\%) & 1\\
MBSTCox & hdfail & 3 (100\%) & 3 (100\%) & 6\\
MBSTAFT & hdfail & 2 (66.7\%) & 0 (0\%) & 2\\
SSVM & check\_times & 3 (100\%) & 0 (0\%) & 3\\
SSVM & child & 3 (100\%) & 0 (0\%) & 3\\
SSVM & colrec & 3 (100\%) & 0 (0\%) & 3\\
SSVM & flchain & 11 (73.3\%) & 0 (0\%) & 11\\
SSVM & hdfail & 3 (100\%) & 0 (0\%) & 3\\
SSVM & nafld1 & 15 (100\%) & 0 (0\%) & 15\\
SSVM & nwtco & 8 (53.3\%) & 0 (0\%) & 8\\
SSVM & ova & 3 (10\%) & 0 (0\%) & 3\\
SSVM & support & 3 (100\%) & 0 (0\%) & 3\\
SSVM & wa\_churn & 15 (100\%) & 0 (0\%) & 15\\
\bottomrule
\end{longtable}

\section{Results per Dataset}\label{app:results_per_dataset}

For completeness, we display boxplots of the evaluation scores across the outer resampling iterations per dataset, learner, evaluation measure and tuning measure.

\begin{figure}[ht]
    \centering
        \includegraphics[width=0.8\textwidth]{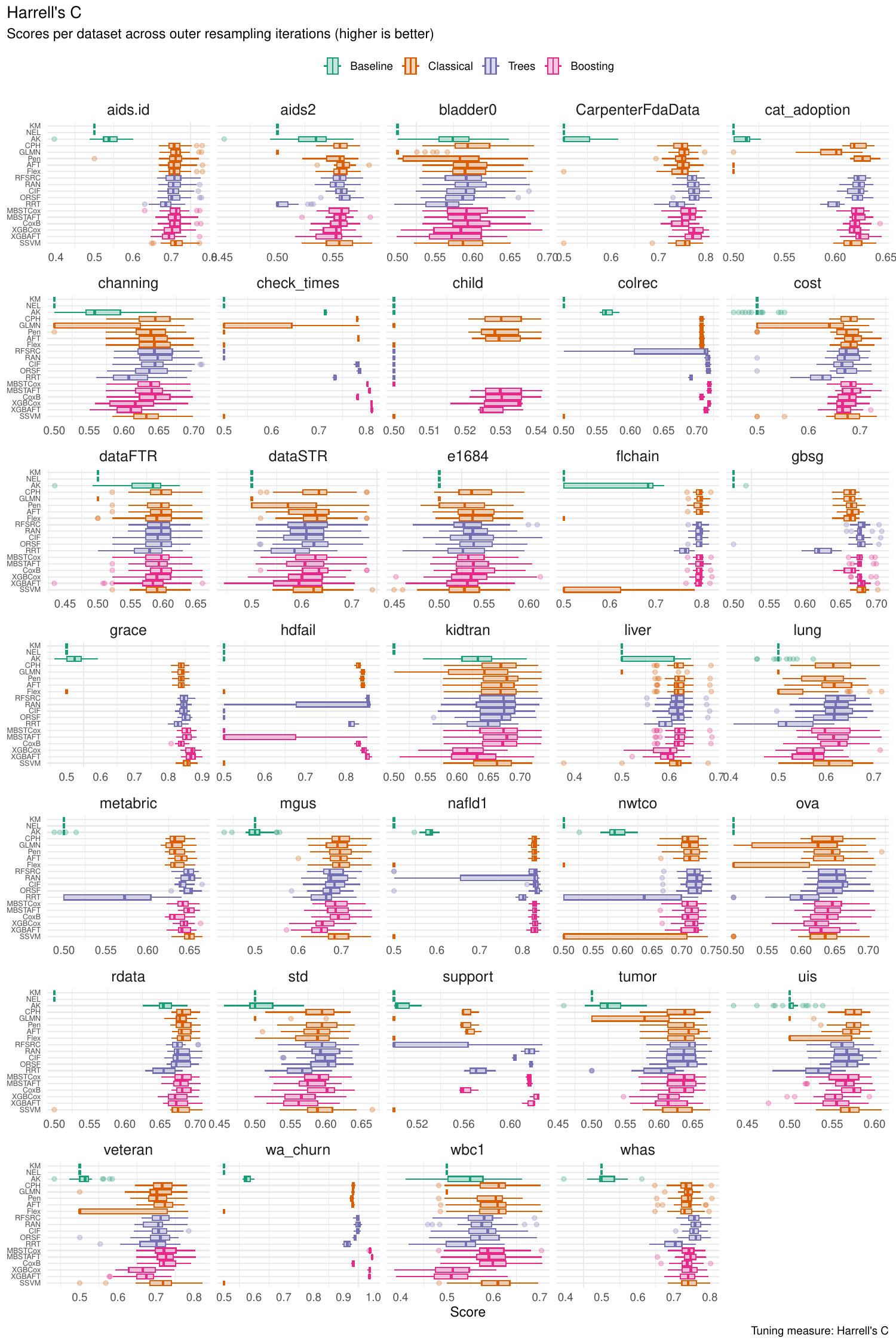}
    \caption{Per-dataset scores for learners tuned and evaluated with Harrell's C}
\end{figure}

\begin{figure}[ht]
    \centering
        \includegraphics[width=0.8\textwidth]{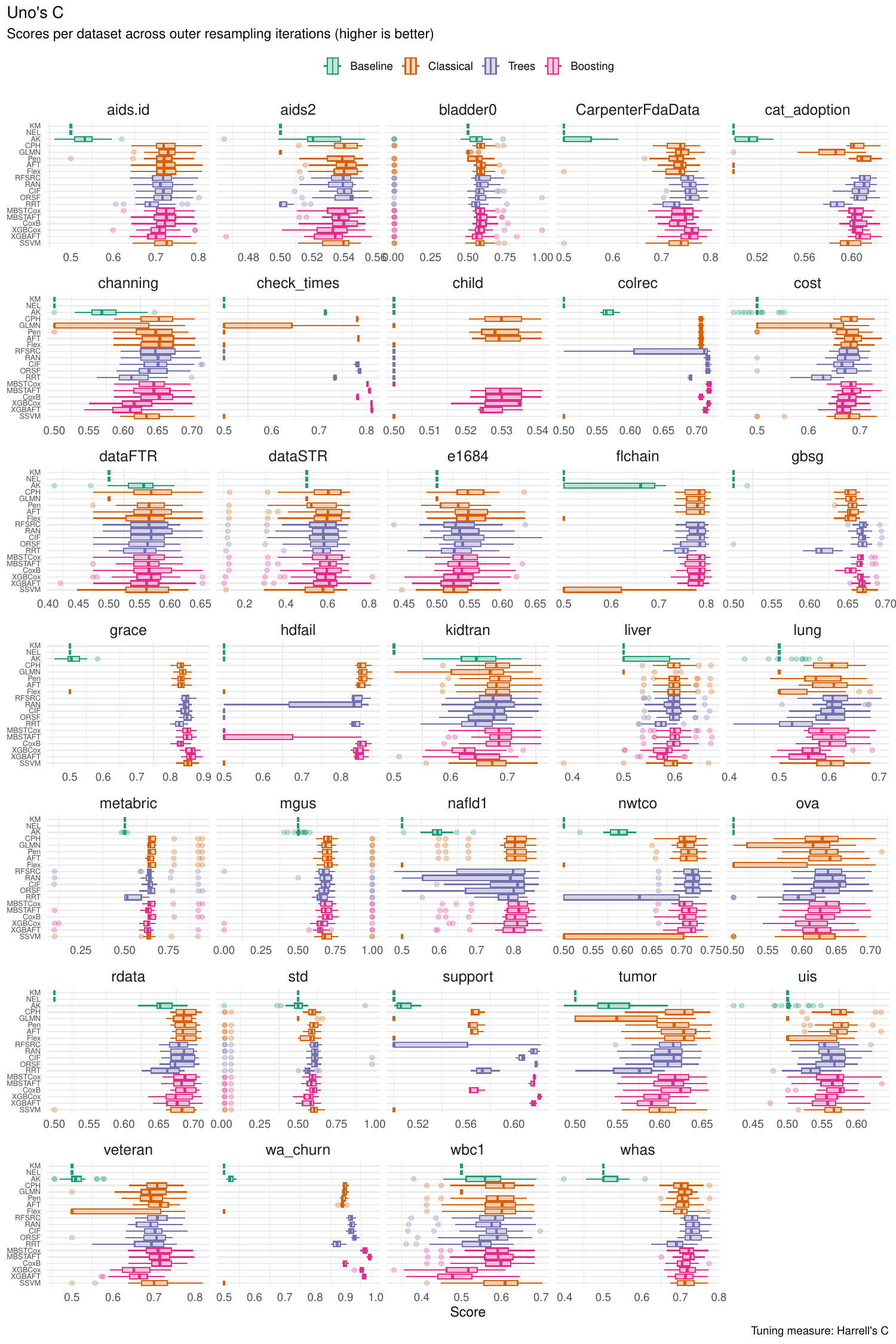}
    \caption{Per-dataset scores for learners tuned on Harrell's C and evaluated with Uno's C}
\end{figure}

\begin{figure}[ht]
    \centering
        \includegraphics[width=0.8\textwidth]{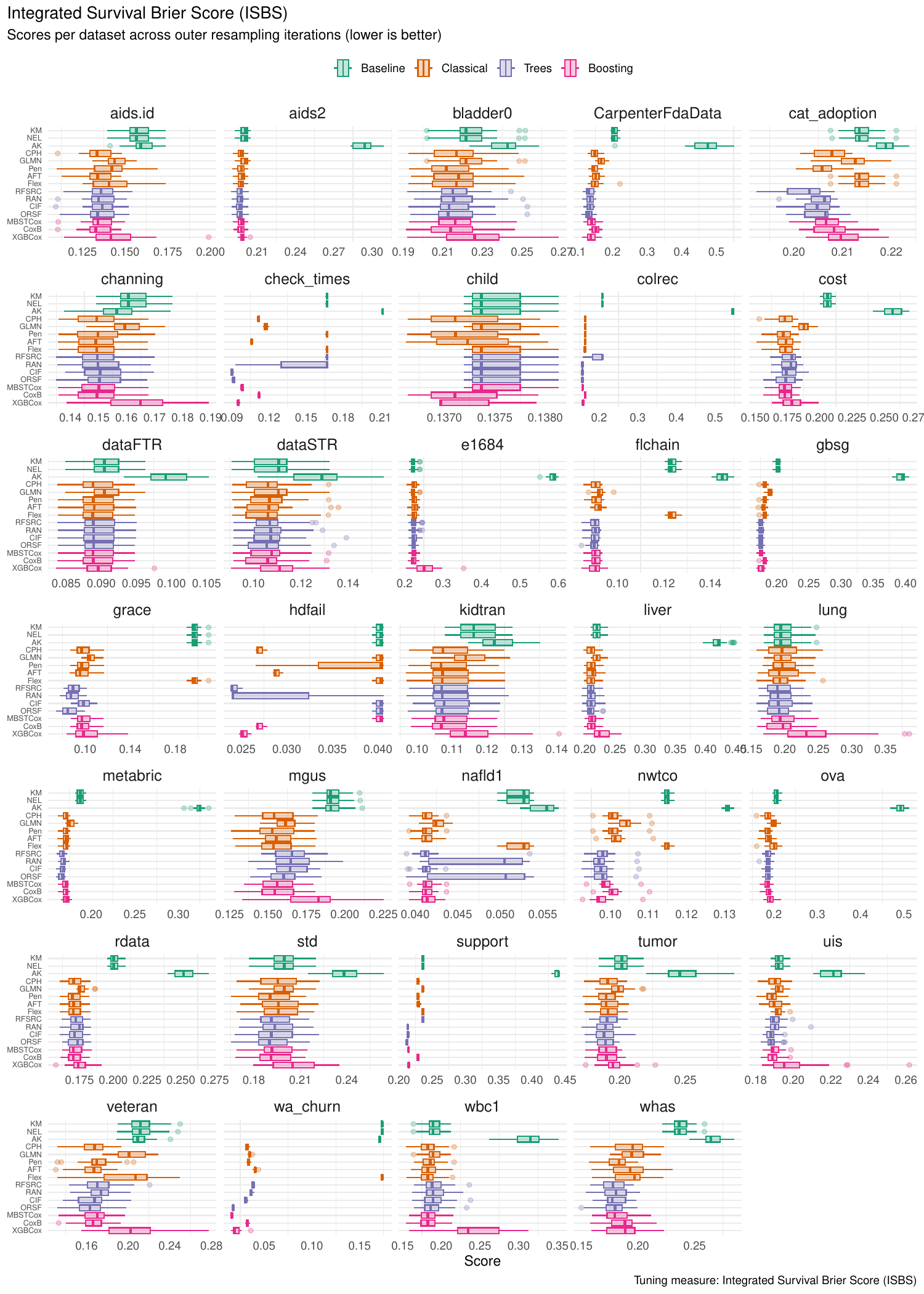 }
    \caption{Per-dataset scores for learners tuned and evaluated with ISBS}
\end{figure}

\begin{figure}[ht]
    \centering
        \includegraphics[width=0.8\textwidth]{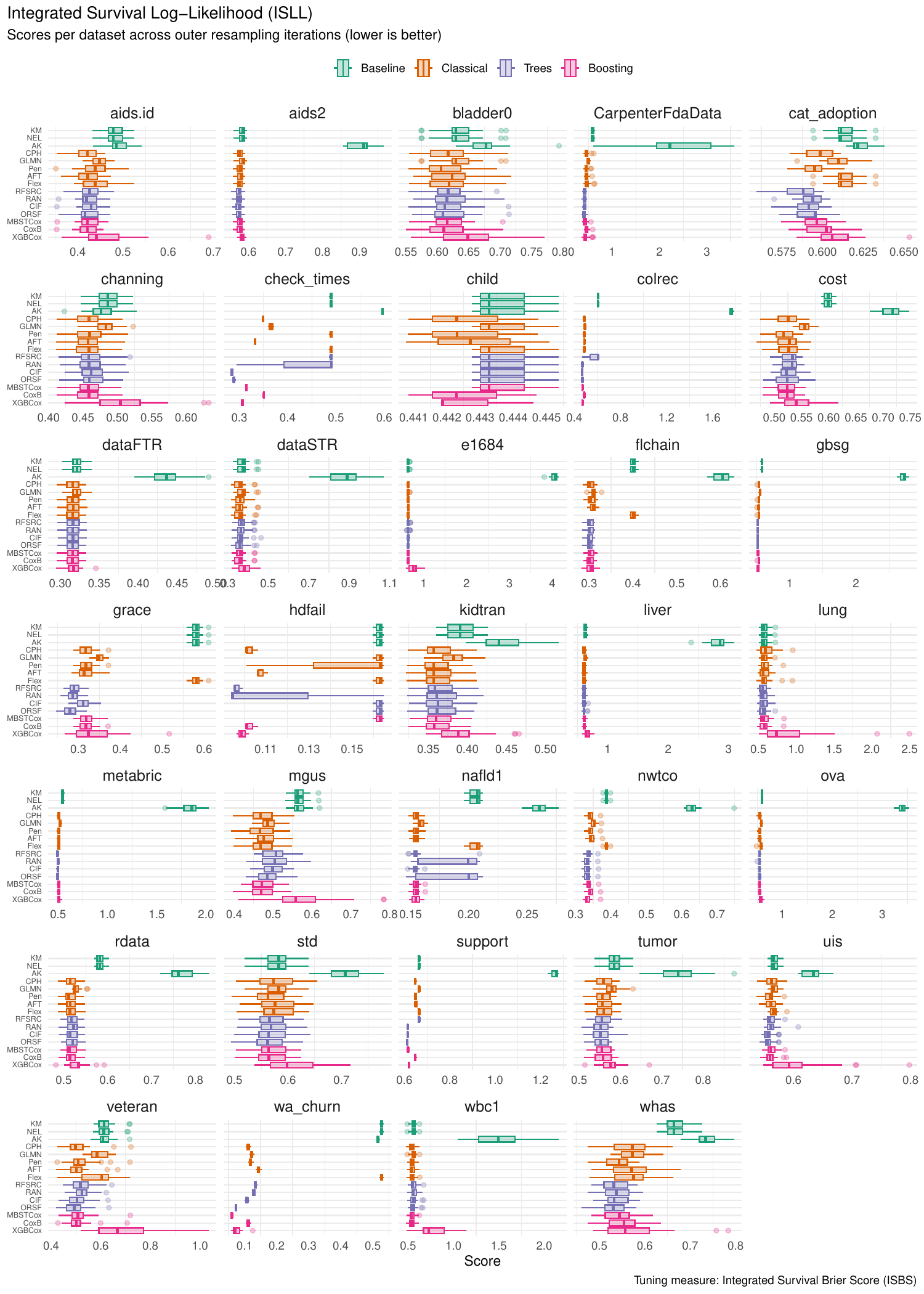}
    \caption{Per-dataset scores for learners tuned on ISBS evaluated with Harrell's C}
\end{figure}
\clearpage

\section{Competing interests}
No competing interest is declared.

\section{Data Availability Statement}
The data underlying this article are available in the online supplement at \href{https://github.com/slds-lmu/paper_2023_survival_benchmark}{\url{https://github.com/slds-lmu/paper_2023_survival_benchmark}}.

\section{Author contributions statement}

LB contributed to the planning and dataset selection, conducted the experiment and wrote the manuscript.
JZ contributed to underlying software implementations and advised on model evaluation.
RS conducted a pilot version of this study \citep{sonabend2021b} which laid the foundation for this project and reviewed the manuscript.
BB, AB, and MNW contributed to the planning, literature review, interpretation, and reviewed the manuscript.

\section{Acknowledgments}

We are grateful for the package maintainers' time supporting this effort regarding the model configurations used in this benchmark, in particular we thank Byron Jaeger, Matthias Schmid, Harald Binder, Cesaire Fouodo, H{\aa}vard Kvamme, Torsten Hothorn, Tong He, Chris Jackson, Terry Therneau, Trevor Hastie, and Jelle Goeman.

This work has been carried out by making use of Wyoming's Advanced Research Computing Center, on its Beartooth Compute Environment (\url{https://doi.org/10.15786/M2FY47}).
We gratefully acknowledge the computational and data resources provided by Wyoming's Advanced Research Computing Center
(\url{https://www.uwyo.edu/arcc/}).

JZ received funding from the European Union's Horizon 2020 research and innovation program under grant agreement No 101016851, project PANCAIM.

BB is supported by the German Research Foundation (DFG), Grant Number: 460135501.

MNW is supported by the German Research Foundation (DFG), Grant Numbers: 437611051, 459360854.

\bibliographystyle{abbrvnat}
\bibliography{reference}

\end{document}